\documentclass{article}
\usepackage{iclr2027_conference,times}
\usepackage{amsmath,amsfonts,bm}

\def\eqref#1{equation~\ref{#1}}
\def\1{\bm{1}}

\DeclareMathAlphabet{\mathsfit}{\encodingdefault}{\sfdefault}{m}{sl}
\SetMathAlphabet{\mathsfit}{bold}{\encodingdefault}{\sfdefault}{bx}{n}

\usepackage{hyperref}
\usepackage{url}
\usepackage{booktabs}
\usepackage{multirow}
\usepackage{array}
\usepackage{graphicx}
\usepackage{amsmath,amssymb}
\usepackage{pifont}
\newcommand{\cmark}{\ding{51}}
\newcommand{\xmark}{\ding{55}}
\usepackage{xcolor}
\usepackage{wrapfig}
\title{\textsc{Sudoku}: Understanding Spatial Organization of Tissue by Learning to Generate}

\author{\mdseries Mo Chen$^{1,2}$, Junjie Pang$^{4}$, Adi Lin$^{3}$, Erpai Luo$^{1}$, Zhen Li$^{1}$, Yuxiang Li$^{3}$, \\
\mdseries Tao Qin$^{2}$, Haiguang Liu$^{2}$, Xuegong Zhang$^{1}$, Shaoshuai Li$^{3*}$, Lei Wei$^{1*}$, Qijin Yin$^{2*}$ \\[5pt]
\mdseries $^{1}$Tsinghua University \quad $^{2}$Zhongguancun Academy \quad $^{3}$BGI Research \\
\mdseries $^{4}$University of the Chinese Academy of Sciences \\[3pt]
\mdseries $^{*}$Corresponding authors}

\iclrfinalcopy
\begin{document}
\maketitle
\lhead{}

\begin{abstract}
Tissue function emerges from the spatial organization of cells across multiple scales. Local interactions shape the state of each cell, and their collective effects give rise to multicellular structures. Spatial transcriptomics (ST) now profiles millions of cells at known coordinates, making it possible to learn this organization from data. Most existing pre-training methods, however, focus on masked reconstruction at fixed masking rates to learn representations. Others generate cells autoregressively along an order that tissue does not have. Here we present \textsc{Sudoku}, a spatial foundation model that learns tissue organization through spatial discrete denoising. \textsc{Sudoku} represents cell states with a hierarchical vocabulary, and learns the distribution of hidden cell states given spatial context across a full masking range. This conditional formulation captures dependencies from local cellular interactions to coherent multicellular configurations without imposing a predefined order, and naturally unifies representation and generation as two inference regimes under full and partial observation. Pre-trained on a rigorously quality-controlled corpus of 49 million cells from 12 organs across human and mouse, \textsc{Sudoku} outperforms existing pre-trained models and specialized methods on niche retrieval, multi-scale clustering, spatial generation and immune-recruitment prediction, taking a step toward a virtual tissue model built from ST data.

\end{abstract}

\section{Introduction}

Tissue function emerges from the spatial organization of cells.
At the local level, the state of a cell is shaped by the identities, states, and
positions of surrounding cells through contact and secreted signals
\citep{ncem,commot}.
Beyond the cellular level, spatially neighboring cells organize into higher-order
tissue structures such as layers, boundaries, and functional niches
\citep{schurch2020,bhate2022}.
Tissue organization therefore reflects a hierarchy of spatial dependencies:
local interactions constrain individual cell states, while their coordinated
effects give rise to collective multicellular organization.
Such structures are not simple aggregates of cells, but emerge from coordinated
spatial relationships among them.
A model of tissue must therefore capture both local cellular dependencies and
the higher-order organization arising from their collective arrangement.

The rapidly growing scale of spatial transcriptomics (ST) makes it increasingly
possible to learn tissue organization directly from data.
Most spatial pre-training methods, however, have focused on representation
learning. They reconstruct masked gene expression or cell states at
one or a few predefined masking levels \citep{cellplm,nicheformer,scgptspatial},
regress latent embeddings \citep{stofm,cellworld,terra}, or use contrastive
objectives \citep{novae} to learn representations of cells and neighborhoods. While representation learning serves important downstream analyses, such objectives expose the model to only one or a limited set of views of spatial dependency.
Yet tissue organization is shaped by complex dependencies spanning individual
cells, local neighborhoods, and collective multicellular configurations.
Learning these dependencies from only a few predefined settings therefore risks
capturing particular spatial relationships rather than the broader organizing
principles of tissue.

Generative modeling offers a complementary perspective by asking which cellular
configurations are compatible with a given spatial context.
GeST \citep{gest}
models this autoregressively by serializing cells into a
one-dimensional sequence. However, a tissue has no natural traversal order, and an imposed one lets a cell condition only on the cells listed before it. 

We argue that tissue organization is best learned by varying the observed spatial context. For a set of spatially neighboring cells with cell states $X$ at coordinates $P$, a general tissue model should learn
\begin{equation}
    p(X_M \mid X_{\bar M}, P),
\end{equation}
for arbitrary hidden subsets $M$.
With only a few cells hidden, the task primarily probes local cell--context
dependencies; as more cells are hidden, it increasingly requires recovering
coherent multicellular configurations.
Varying the amount of observed context therefore exposes the model to a continuum
from local cellular constraints to collective tissue organization. A model trained on this conditional then needs no separate  objective for representation downstream tasks, as effective generative pre-training has been shown to yield transferable representations in language \citep{gpt1,mdlm} and vision \citep{igpt,mage,visionbanana}.

\begin{figure}[t]
\centering
\includegraphics[width=\linewidth]{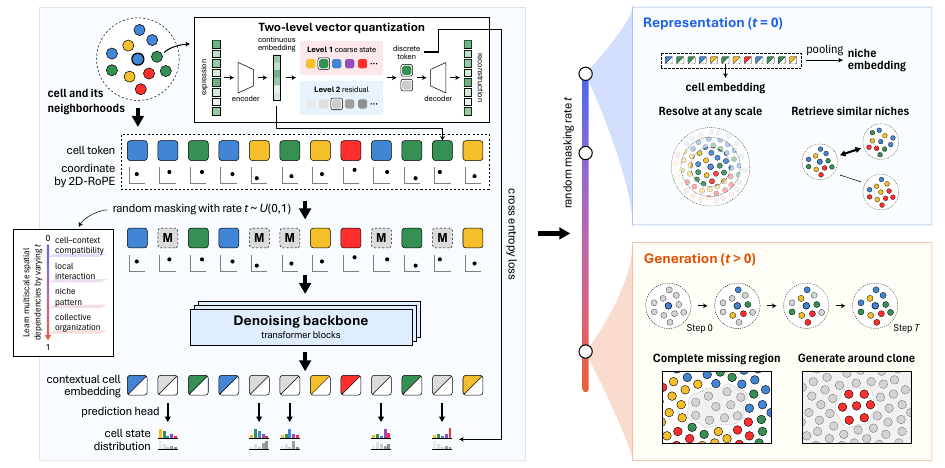}
\caption{
Spatial discrete denoising.
A coordinate-aware spatial transformer predicts masked cellular states from
visible spatial context across variable masking rates, learning dependencies from
local cell--context relationships to multicellular organization.
The same model provides contextual representations under full observation and
generates missing cellular configurations under partial observation.
}
\label{fig:overview}
\end{figure}

Motivated by this view, we propose \textbf{\textsc{Sudoku}}\footnote{Named after the puzzle, in which each blank cell is filled with one of a fixed set of symbols, constrained by the filled cells around it.}, a spatial foundation model
trained with \textbf{spatial discrete denoising} (Figure \ref{fig:overview}).
A residual-quantized tokenizer maps whole-cell transcriptomes into a hierarchical
vocabulary of discrete cell states, and a coordinate-aware spatial transformer
is trained to recover masked cells from their visible context.
By sampling the masking rate across nearly the full interval from zero to one,
\textsc{Sudoku} learns spatial dependencies under continuously varying amounts of
observed context, from local cell--context relationships to coherent
multicellular organization, without imposing an artificial order on cells. Our main contributions are:

(1) We formulate spatial tissue pre-training as learning conditional cellular distributions under arbitrary partial observation, providing a unified view of local cellular dependence and collective tissue organization.

(2) We introduce spatial discrete denoising as a new pre-training objective for ST. This approach combines a hierarchical cell-state vocabulary with a bidirectional, order-free spatial transformer trained across the full masking-rate spectrum. The pre-trained backbone serves as an encoder when the neighborhood is fully observed and performs generation when there are missing regions.
    
(3) We pre-train \textsc{Sudoku} on a high-quality corpus of 49 million cells from 12 organs across human and mouse, and show broad transfer across both representation and generation, outperforming existing methods on niche retrieval, multi-scale clustering, spatial generation, and immune-recruitment prediction.

\section{Background and related work}

\subsection{Key properties of ST data and challenges of data modeling}
\label{sec:data}

ST profiles each tissue section as a gene expression matrix, where each row is a cell and each column one of about twenty thousand genes, along with an additional two-dimensional physical coordinate for every cell. Here we show three key properties of such data, which bring challenges and call for specialized design of the training objective and model architecture. 

\paragraph{Expression is high-dimensional and noisy, and cell states are hierarchical.} Gene-expression profiles are high-dimensional, sparse, and noisy: most entries of the expression matrix are zero \citep{svensson2020,xtrimogene}, while observed counts also reflect technical artifacts \citep{hafemeister2019,hartman2025}. A neighborhood of hundreds of cells amounts to millions of measurements, too long to model as gene sequences. Meanwhile, cell states are hierarchically organized, from broad cell types to finer subtypes and functional states \citep{zeng2022,domcke2023,bian2025}. Predicting cell states as a flat set of unrelated categories thus incorrectly penalizes a closely related state as heavily as an unrelated one.

\paragraph{Spatial dependence is both local and collective.} Cell states are shaped by the identities and positions of surrounding cells through contact and secreted signals \citep{ncem,commot}. These local relationships collectively organize neighboring cells into recurrent multicellular structures that define higher-order tissue organization \citep{schurch2020,bhate2022}. A model trained to predict limited missing cells may learn how each cell depends on its surroundings, but can hardly understand how cells are jointly arranged into these structures.

\paragraph{Tissues are not naturally one-dimensional.} A tissue section is a set of cells embedded in physical space. Each cell can be jointly influenced by neighbors from all directions \citep{ncem,commot,spatialdm,spatalk}, and their relative spatial position does not change with how cells are listed. Therefore, serializing tissue into one-dimensional cell sequences for autoregressive generation imposes an artificial dependency structure, in which each cell is conditioned only on the cells placed before it, although the neighbors after it influence it just as much.

Together, these properties require a tissue model to represent cellular states compactly, capture spatial dependencies from local cell–context relationships to collective organization, and preserve the native two-dimensional structure of tissue.

\subsection{Existing pre-training paradigms for ST}

\textbf{Representation learning.}
Most existing methods are designed to learn transferable representations of cells or their spatial context. Within this paradigm, several types of pre-training objectives have been
explored.

\textit{Masked reconstruction in expression space} predicts the expression of masked genes or cells from the remaining context. CellPLM \citep{cellplm}, Nicheformer \citep{nicheformer} and scGPT-spatial \citep{scgptspatial} adopt this strategy with one or a few fixed masking rates. While CellPLM and scGPT-spatial are able to impute missing gene expression profiles, their capability is limited to imputation of partly observed cells rather than generating completely missing cell states. 

\textit{Masked reconstruction in latent space} moves the reconstruction target to a cell embedding. SToFM \citep{stofm} regresses the embeddings of masked cells from a single-cell foundation model, and CellWorld \citep{cellworld} and TERRA \citep{terra} predict the latent from a teacher encoder. While this keeps prediction in a compact latent space with less noise, the output is still a point estimate that cannot be sampled.

\textit{Contrastive learning} avoids explicit reconstruction. Novae learns neighborhood representations through graph contrastive learning and prototype assignment \citep{novae}. These methods are common for representation learning, but do not explicitly model the distribution of cellular configurations conditioned on spatial context.

\textbf{Generative modeling.}
Generative models instead aim to learn a distribution over cellular states or
configurations. GeST \citep{gest} represents cells with discrete meta cells obtained from pre-clustering and generates them autoregressively. To apply autoregression, however, cells must first be serialized into a one-dimensional sequence with an imposed traversal order.

Table~\ref{tab:methods} summarizes these spatial pre-training paradigms and
their differences in prediction target, masking strategy, generative capability,
and dependence on cell ordering.
\begin{table}[t]
\centering
\caption{Pre-training objectives of spatial transcriptomics models. Mask rate: masking rate during training; Generative: whether the output is a distribution over cell states that can be sampled; Order-free: whether generation imposes no order on the cells; Multi-scale: whether the pre-trained model gives representations at more than one spatial scale.}
\label{tab:methods}
\footnotesize
\setlength{\tabcolsep}{1.5pt}
\begin{tabular}{l l l l c c c}
\toprule
Method          & Token & Pre-training task            & Mask rate              & Generative & Order-free & Multi-scale \\
\midrule
CellPLM         & cell  & masked expression regression & 0.75 cells, 0.25 genes & \xmark     & --         & \xmark      \\
Nicheformer     & gene  & masked gene prediction       & 0.15                   & \xmark     & --         & \xmark      \\
scGPT-spatial   & gene  & masked expression regression & 0.25 / 0.5 / 0.75      & \xmark     & --         & \xmark      \\
SToFM           & cell  & masked embedding regression  & 0.10                   & \xmark     & --         & \cmark      \\
CellWorld       & cell  & masked latent regression     & 0.6                    & \xmark     & --         & \xmark      \\
TERRA           & gene  & masked latent regression     & 0.6                    & \xmark     & --         & \xmark      \\
Novae           & --    & contrastive learning         & --                     & \xmark     & --         & \cmark      \\
GeST            & cell  & autoregressive generation    & --                     & \cmark     & \xmark     & \xmark      \\
\midrule
\textsc{Sudoku} & cell  & discrete denoising           & $U(0,1)$               & \cmark     & \cmark     & \cmark      \\
\bottomrule
\end{tabular}
\end{table}

\subsection{Masked discrete generative modeling}
\label{sec:discrete_generation}

Before spatial transcriptomics, masked discrete modeling had proved effective as an alternative generative paradigm to autoregressive factorization in other data modalities, including vision and natural language. D3PM introduced diffusion processes in discrete state spaces \citep{d3pm}, while MaskGIT showed that masked tokens can be generated
iteratively by repeatedly predicting and committing high-confidence states
\citep{maskgit}. More recently, MDLM and RADD established that variable-rate masked prediction corresponds to the variational objective of an absorbing-state discrete diffusion process \citep{mdlm,radd}, and LLaDA demonstrated that this formulation can scale to large language models \citep{llada}.

\textsc{Sudoku} builds on this line of work but differs in several aspects that
follow from the properties of ST data in Section~\ref{sec:data}: what a token is,
how tokens are placed in space, and how the visible context enters the model.

\section{Method}
\label{sec:method}

A tissue section is a set of $n$ cells, $S=\{(\mathbf{p}_i,\mathbf{x}_i)\}_{i=1}^{n}$, where $\mathbf{p}_i\in\mathbb{R}^2$ is the physical coordinate of cell $i$ on the section and $\mathbf{x}_i\in\mathbb{R}^{G}$ is its expression vector over $G$ genes. The neighborhood $N_i$ of cell $i$ is the set of cells within radius $r$ of it. The task is to infer the states of the cells hidden in a neighborhood from the states of the visible ones and the coordinates of all of them.

\textsc{Sudoku} consists of two components: a hierarchical tokenizer that maps continuous cellular profiles into discrete cell states, and a coordinate-aware spatial transformer trained by spatial discrete denoising. Together, they define this conditional distribution.

\subsection{Hierarchical cell-state tokenizer}
\label{sec:tokenizer}

Discrete denoising requires a compact state space over which masked cellular states can be predicted probabilistically. We therefore encode each cell into a discrete representation of its transcriptional state, reducing a high-dimensional expression profile to a small number of cell-state codes while preserving relationships among related states. This also makes a neighborhood of hundreds of cells tractable as a collection of cellular states rather than millions of gene-level measurements.

Because cellular states are hierarchically organized, we use residual quantization rather than a flat codebook. The first-level code captures a coarse state and the second-level code refines the residual variation within it. This structure makes nearby cellular states partially share their discrete representation rather than treating all state mismatches equally.

The tokenizer maps the expression $\mathbf{x}_i\in\mathbb{R}^{G}$ of a cell over $G$ genes to a two-level residual-quantized code. The encoder $f_\varphi$ is an MLP over normalized log expression that outputs a 128-dimensional latent $\mathbf{z}_i=f_\varphi(\mathbf{x}_i)$. Two codebooks $\mathbf{e}^{(1)},\mathbf{e}^{(2)}$ each hold $K=256$ vectors of dimension 128, making the vocabulary size $K^2=65{,}536$. The first level takes the code nearest to $\mathbf{z}_i$, the second level the code nearest to the residual, and the quantized latent is their sum:
\begin{equation}
c^{(1)}_i=\mathrm{NN}\big(\mathbf{z}_i;\mathbf{e}^{(1)}\big),\qquad
c^{(2)}_i=\mathrm{NN}\big(\mathbf{z}_i-\mathbf{e}^{(1)}_{c^{(1)}_i};\mathbf{e}^{(2)}\big),\qquad
\mathbf{z}_{q,i}=\mathbf{e}^{(1)}_{c^{(1)}_i}+\mathbf{e}^{(2)}_{c^{(2)}_i},
\end{equation}
where $\mathrm{NN}(\cdot;\mathbf{e})$ returns the index of the nearest code.
A decoder $g_\varphi$ maps $\mathbf{z}_{q,i}$ to the mean and dispersion of a negative-binomial distribution per gene \citep{scvi}, $(\boldsymbol{\mu}_i,\mathbf{r}_i)=g_\varphi(\mathbf{z}_{q,i})$, with the mean scaled by the total count $s_i$ of the cell:
\begin{equation}
\mathrm{NB}\big(x_{ig}\mid s_i\mu_{ig},r_{ig}\big)=\frac{\Gamma(x_{ig}+r_{ig})}{\Gamma(r_{ig})\,\Gamma(x_{ig}+1)}\Big(\frac{r_{ig}}{r_{ig}+s_i\mu_{ig}}\Big)^{r_{ig}}\Big(\frac{s_i\mu_{ig}}{r_{ig}+s_i\mu_{ig}}\Big)^{x_{ig}}.
\end{equation}
The training loss is the negative log-likelihood summed over genes plus a commitment term with $\beta=1$,
\begin{equation}
\mathcal{L}_{\mathrm{tok}}(\varphi)=-\sum_{g=1}^{G}\log\mathrm{NB}\big(x_{ig}\mid s_i\mu_{ig},r_{ig}\big)+\beta\,\big\|\mathbf{z}_i-\mathrm{sg}(\mathbf{z}_{q,i})\big\|_2^2,
\end{equation}
where $\mathrm{sg}$ is the stop-gradient. Codebooks track the exponential moving average of the latents assigned to each code. The joint code is written $c_i=(c^{(1)}_i,c^{(2)}_i)$. The distance used at each quantization level, network widths, moving-average decay and batch composition are in Appendix~\ref{app:tokenizer}.

\subsection{Spatial discrete denoising}
\label{sec:pretraining}

We now train the spatial transformer to recover arbitrary subsets of cellular states from the remaining spatial context. By varying the masking rate across neighborhoods, the model is exposed to conditional inference problems ranging from local cell-state recovery to reconstruction of largely missing multicellular configurations.

\paragraph{Forward process.} The forward process corrupts a neighborhood by masking part of its cells, and the spatial transformer is trained to reverse it. For each neighborhood $N_i$ a masking rate $t\sim U(\epsilon,1)$ is drawn independently, with $\epsilon=0.001$, so that the model learns to generate from any amount of visible context. Each cell is masked independently with probability $t$, and both levels of a masked cell's code are replaced by the mask symbol $[\mathrm{M}]$:
\begin{equation}
m_j\sim\mathrm{Bernoulli}(t),\qquad M=\{\,j\in N_i: m_j=1\,\},\qquad V=N_i\setminus M.
\end{equation}

\paragraph{Input.} The input to the spatial transformer consists of gene expression information and spatial information. For gene expression, visible cells are fed the linear projection $\mathbf{W}\mathbf{z}_k$ of their continuous tokenizer latent and masked cells a learned mask vector $\mathbf{u}$, and together they form the layer-0 input $\mathbf{h}^0_j$. For spatial information, coordinates are first divided by the measured median nearest-neighbor spacing $d_s$ of section $s$, $\hat{\mathbf{p}}_j=\mathbf{p}_j/d_s$, to align physical scale across platforms. Then $\hat{\mathbf{p}}$ enters every attention layer through an axial rotary positional encoding, in which the head dimension $D$ is split in half, the first half rotated by the $x$ component of $\hat{\mathbf{p}}$ and the second by the $y$ component, with frequency $\omega_m=\theta^{-2m/(D/2)}$ for the $m$-th pair of dimensions and $\theta=10$. Writing $\mathbf{R}(\hat{\mathbf{p}})=\mathrm{diag}(\mathbf{R}_x(\hat{p}_x),\mathbf{R}_y(\hat{p}_y))$ for the rotation and $\mathbf{q}_j,\mathbf{k}_k$ for the query of cell $j$ and the key of cell $k$ at a layer, the attention score is
\begin{equation}
a_{jk}=\big(\mathbf{R}(\hat{\mathbf{p}}_j)\mathbf{q}_j\big)^{\top}\big(\mathbf{R}(\hat{\mathbf{p}}_k)\mathbf{k}_k\big)=\mathbf{q}_j^{\top}\mathbf{R}(\hat{\mathbf{p}}_k-\hat{\mathbf{p}}_j)\,\mathbf{k}_k,
\end{equation}
so the score depends only on the relative displacement of the two cells; we ablate the positional encoding in Table~\ref{tab:ablation}. Theoretically, this way of positional encoding is invariant to translation but not to rotation. But in practice, we show that rotating the coordinates has almost no impact on the model (Appendix~\ref{app:rotation}). Continuous input with discrete targets follows BEiT v2 \citep{beitv2}. This keeps more information of the visible cells and asks the model to learn the conditional cell state distribution at missing positions (Table~\ref{tab:ablation}).

\paragraph{Backbone and output.} The backbone is a transformer with 10 layers, width 256 and 8 heads, and the final hidden state of cell $j$ is $\mathbf{h}_j$. Two classification heads give the $K$-way distributions of the first- and second-level codes, $p_\theta(c^{(\ell)}_j\mid\mathbf{z}_V,\mathbf{p}_N)$ for $\ell=1,2$, where $p_\theta$ denotes the spatial transformer, $\mathbf{z}_V$ are the latents of the visible cells and $\mathbf{p}_N$ the coordinates of all cells in the neighborhood. Within a single forward pass, the code levels and masked cells are factorized conditionally on the visible context; we explain this design choice in Appendix~\ref{app:tokenizer}.

\paragraph{Loss.} The cross-entropy of both code levels on the masked cells, weighted by $1/t$:
\begin{equation}
\mathcal{L}(\theta)=\mathbb{E}_{t}\,\mathbb{E}_{M\sim\mathrm{Bern}(t)}\ \frac{1}{t}\sum_{j\in M}\sum_{\ell=1}^{2}-\log p_\theta\big(c^{(\ell)}_j\,\big|\,\mathbf{z}_V,\mathbf{p}_N\big).
\label{eq:loss}
\end{equation}
With the $1/t$ weighting, Equation~\ref{eq:loss} is the variational bound of absorbing-state discrete diffusion on the codes of a neighborhood \citep{mdlm,radd}. Minimizing it therefore trains a generative model of the whole neighborhood. Predicting a distribution over codes, rather than regressing the latent of a masked cell, has two benefits. A distribution can be sampled while a point estimate cannot. Regression is optimized by the average cell state given the context, whereas the distribution also separates contexts that admit different cell states with the same average, so the representations of the visible cells must encode more about the context. Replacing it by regression lowers both representation and generation (Table~\ref{tab:ablation}, Appendix~\ref{app:ablation}). Note that visible cells enter through their continuous latents $\mathbf{z}_V$ rather than through their codes, different from the canonical MDLM objective \citep{mdlm}. We show in Appendix~\ref{app:bound} that the bound is unchanged, since the latent determines the code, thus the generative modeling property is preserved. The $1/t$ weighting balances contributions from different masking levels, so the model learns conditional distributions across nearly the full range of observed context, from almost complete neighborhoods to largely missing configurations. The objective factorizes over the masked subset and needs no traversal order.

\paragraph{Pre-training corpus.} We curate a high-quality pre-training corpus of 396 single-cell resolution whole-transcriptome tissue sections, 161 human and 235 mouse, from 12 organs and from both normal and tumor tissue, 49.1 million cells in all, filtered by per-cell and per-section quality control. The corpus composition, quality control and hyperparameters are in Appendix~\ref{app:tokenizer}.

\subsection{Inference under partial and full observation}
\label{sec:inference}

\paragraph{Generation.} Starting from the missing set fully masked, unmask in $T$ parallel steps. At step $s$ let $M_s$ be the set still masked, with $M_1=M$. One forward pass gives the two-level distribution of every $j\in M_s$. Each level's code is drawn by the Gumbel-max trick from the distribution tempered by a temperature $\tau$, and the confidence of a cell is the probability of its drawn codes,
\begin{equation}
\hat{c}^{(\ell)}_j=\arg\max_{k}\Big[\tfrac{1}{\tau}\log p_\theta\big(c^{(\ell)}_j=k\,\big|\,\cdot\big)+g_{jk}\Big],\qquad
\sigma_j=\sum_{\ell=1}^{2}\log p_\theta\big(\hat{c}^{(\ell)}_j\,\big|\,\cdot\big),
\end{equation}
with $g_{jk}\sim\mathrm{Gumbel}(0,1)$ drawn independently. The $k_s$ cells with the highest $\sigma_j$ are committed with their drawn codes. At the next step they enter the conditioning through their codebook reconstruction $\mathbf{z}_q(\hat{c}_j)$${}=\mathbf{e}^{(1)}_{\hat{c}^{(1)}_j}+\mathbf{e}^{(2)}_{\hat{c}^{(2)}_j}$, and the rest remain in $M_{s+1}$. The commit count is near-uniform with the remainder placed in the last steps, $k_s=\lfloor|M|/T\rfloor$ plus one when $s>T-(|M|\bmod T)$, so that the hardest cells are decided with the most context. At $\tau=1$ every committed code is a sample from $p_\theta$, and as $\tau\to0$ the draw becomes the argmax and decoding is greedy. Within a step the committed cells are drawn independently given the current context, while dependence between generated cells is carried across steps, since every committed cell conditions all later draws, so the $T$ steps carry dependence across generated cells and approximate a sample of the joint distribution over $M$.

\paragraph{Representation.} When performing representation tasks, the neighborhood is fully visible. The final hidden state $\mathbf{h}_i$ of the center cell $i$ is its representation in context, and the neighborhood representation is the mean of hidden states within it,
\begin{equation}
\bar{\mathbf{h}}_i=\frac{1}{|N_i|}\sum_{k\in N_i}\mathbf{h}_k .
\end{equation}
The neighborhood radius $r$ controls the spatial scale of the readout and can be varied to obtain representations at multiple levels of tissue organization.

\section{Experiments}
\label{sec:experiments}

We evaluate \textsc{Sudoku} on five benchmarks that cover the two regimes of Section~\ref{sec:inference}. Under full observation, we evaluate \textsc{Sudoku}'s representation quality with niche retrieval on the human embryo atlas HESTA \citep{hesta} and the human cortex dataset DLPFC \citep{dlpfc}, and multi-scale clustering on MERFISH mouse brain data \citep{wtmerfish}. Under partial observation, we test \textsc{Sudoku}'s spatial generation on in-house sections of human and mouse brain and mouse kidney. Furthermore, we test \textsc{Sudoku}'s generative ability in modeling tumor microenvironment and predicting immune recruitment  of tumor clones in SPAC-seq, a spatial CRISPR screen of mouse lung metastases \citep{spacseq} (Section~\ref{sec:spac}). All benchmark sections are held out of pre-training. More details of datasets and evaluation protocols are in Appendix~\ref{app:benchmarks} and Appendix~\ref{app:baselines}. We report results of \textsc{Sudoku} without fine-tuning in Appendix~\ref{app:zeroshot}.

\subsection{Representing tissue organization across sections and scales}
\label{sec:representation}

\paragraph{Niche retrieval.} We first test whether \textsc{Sudoku}'s representations recognize the same niche, a spatially localized group of cells with certain features, across independent tissue sections. Given a niche of interest (NOI) in a query section, the task is to retrieve the matching niches from other sections by the similarity of their representations, an operation behind atlas query and label transfer. HESTA provides organ-level NOIs from human embryo Stereo-seq sections at single-cell resolution, scored by cross-section retrieval AUPRC. DLPFC provides finer NOIs at cortical-layer junctions from 12 human cortex Visium sections at spot resolution, a platform absent from pre-training, scored by PCC against a matching score computed from the layer labels. The task, the metrics and intuitive visualizations are in Appendix~\ref{app:benchmarks}.

On HESTA, \textsc{Sudoku} achieves the highest AUPRC on seven of the eight organs, with GeST leading only on jaw and teeth. Notably, \textsc{Sudoku} outperforms the specialized niche-retrieval method QueST \citep{quest} on all eight organs (Table~\ref{tab:retrieval}). We report the chance level of each organ, chance-normalized scores and the standard deviations over three seeds in Appendix~\ref{app:retrieval-extra}. On DLPFC, \textsc{Sudoku} achieves the highest PCC on five of the seven niche classes, ties QueST on one, and is second only to QueST on the remaining class. These results show that the representations of \textsc{Sudoku} support niche retrieval at both the organ and the layer-junction level, on single-cell data and on spot-level data from a platform unseen in pre-training.

\begin{table}[htb]
\centering
\caption{Niche retrieval on HESTA, AUPRC per organ, and DLPFC, PCC per NOI type (the junction of two or three cortical layers including L3, L4, L5, L6 and White Matter). Rows are grouped into task-specific methods, spatial pre-trained models and ours. \textbf{Bold}: best; \underline{underline}: second best.}
\label{tab:retrieval}
\scriptsize
\setlength{\tabcolsep}{1.5pt}
\begin{tabular*}{\linewidth}{@{\extracolsep{\fill}} l cccccccc ccccccc @{}}
\toprule
& \multicolumn{8}{c}{HESTA, AUPRC per organ} & \multicolumn{7}{c}{DLPFC, PCC per neighborhood class} \\
\cmidrule(lr){2-9}\cmidrule(lr){10-16}
Method          & Brain             & Liver             & Spinal            & Cartilage         & Ganglion          & Muscle            & Heart             & Jaw               & L3L4           & L4L5              & L5L6              & L6WM              & L3L4L5            & L4L5L6            & L5L6WM            \\
\midrule
QueST           & \underline{0.490} & 0.797             & \underline{0.430} & \underline{0.352} & 0.147             & 0.277             & \underline{0.710} & 0.126             & \textbf{0.566} & \underline{0.673} & \underline{0.762} & \underline{0.746} & \textbf{0.628}    & \underline{0.720} & \underline{0.696} \\
\midrule
CellPLM         & 0.323             & 0.731             & 0.303             & 0.278             & 0.129             & 0.276             & 0.471             & 0.148             & 0.344          & 0.285             & 0.238             & 0.015             & 0.424             & 0.269             & 0.126             \\
Novae           & 0.423             & 0.333             & 0.396             & 0.238             & 0.073             & 0.147             & 0.101             & 0.162             & 0.508          & 0.603             & 0.511             & 0.620             & 0.580             & 0.557             & 0.479             \\
Nicheformer     & 0.334             & 0.589             & 0.350             & 0.268             & 0.121             & 0.166             & 0.208             & 0.136             & 0.265          & 0.240             & 0.120             & 0.009             & 0.352             & 0.213             & $-$0.005          \\
scGPT-spatial   & 0.328             & 0.624             & 0.302             & 0.250             & 0.111             & 0.270             & 0.391             & 0.149             & 0.312          & 0.212             & 0.123             & $-$0.005          & 0.368             & 0.191             & 0.013             \\
SToFM           & 0.228             & 0.467             & 0.234             & 0.214             & 0.075             & 0.121             & 0.063             & 0.098             & 0.073          & 0.128             & 0.135             & 0.123             & 0.120             & 0.124             & 0.112             \\
CellWorld       & 0.224             & 0.574             & 0.207             & 0.159             & 0.082             & 0.135             & 0.081             & 0.086             & 0.261          & 0.325             & 0.222             & $-$0.094          & 0.382             & 0.305             & 0.059             \\
TERRA           & 0.387             & 0.532             & 0.381             & 0.235             & 0.154             & 0.186             & 0.190             & 0.155             & 0.326          & 0.282             & 0.214             & $-$0.030          & 0.409             & 0.261             & 0.096             \\
GeST            & 0.446             & \underline{0.847} & 0.395             & 0.315             & \underline{0.251} & \underline{0.410} & 0.657             & \textbf{0.253}    & 0.465          & 0.629             & 0.680             & 0.626             & 0.539             & 0.646             & 0.606             \\
\midrule
\textsc{Sudoku} & \textbf{0.517}    & \textbf{0.932}    & \textbf{0.503}    & \textbf{0.359}    & \textbf{0.264}    & \textbf{0.427}    & \textbf{0.758}    & \underline{0.207} & \textbf{0.566} & \textbf{0.698}    & \textbf{0.765}    & \textbf{0.751}    & \underline{0.604} & \textbf{0.723}    & \textbf{0.705}    \\
\bottomrule
\end{tabular*}
\end{table}

\begin{table}[t]
\centering
\caption{Multi-scale clustering on mouse brain, median ARI and NMI at three levels: Region (Allen brain division), Band (laminar band) and Cell type. Rows: task-specific methods, pre-trained models, ours. \textbf{Bold}: best; \underline{underline}: second best.}
\label{tab:multiscale}
\footnotesize
\renewcommand{\arraystretch}{0.9}
\begin{tabular*}{\linewidth}{@{\extracolsep{\fill}} l cc cc cc @{}}
\toprule
& \multicolumn{2}{c}{Region} & \multicolumn{2}{c}{Band} & \multicolumn{2}{c}{Cell type} \\
\cmidrule(lr){2-3}\cmidrule(lr){4-5}\cmidrule(lr){6-7}
Method          & ARI               & NMI               & ARI               & NMI               & ARI               & NMI               \\
\midrule
STAGATE         & \underline{0.312} & \underline{0.469} & \underline{0.420} & \textbf{0.634}    & 0.154             & 0.383             \\
BANKSY          & 0.111             & 0.190             & 0.091             & 0.186             & \underline{0.440} & 0.630             \\
\midrule
CellPLM         & 0.113             & 0.170             & 0.115             & 0.140             & 0.438             & \underline{0.671} \\
Novae           & 0.264             & 0.440             & 0.334             & 0.535             & 0.121             & 0.345             \\
Nicheformer     & 0.246             & 0.294             & 0.144             & 0.180             & 0.345             & 0.567             \\
scGPT-spatial   & 0.177             & 0.235             & 0.223             & 0.253             & 0.364             & 0.605             \\
SToFM           & 0.176             & 0.237             & 0.229             & 0.357             & 0.283             & 0.494             \\
CellWorld       & 0.232             & 0.309             & 0.026             & 0.043             & 0.267             & 0.435             \\
TERRA           & 0.207             & 0.304             & 0.212             & 0.355             & 0.154             & 0.390             \\
\midrule
\textsc{Sudoku} & \textbf{0.342}    & \textbf{0.491}    & \textbf{0.493}    & \underline{0.629} & \textbf{0.520}    & \textbf{0.776}    \\
\bottomrule
\end{tabular*}
\end{table}

\paragraph{Multi-scale clustering.} We next ask whether the same backbone resolves tissue at multiple levels via representations at different spatial scales. We use six whole-transcriptome MERFISH sections of mouse brain annotated at three levels: cell type, cortical laminar band and Allen brain division \citep{allenccf}. For \textsc{Sudoku}, cell-type clustering uses the contextual cell representation within 50~$\mu$m, while laminar and regional clustering use pooled neighborhood representations at 50 and 100~$\mu$m, respectively. For the other methods, we read out one embedding per cell and pool it over the same radii. Clustering performance is evaluated by ARI and NMI.

\textsc{Sudoku} achieves the highest ARI at all three organizational levels and the highest NMI at the cell-type and regional levels, and nearly ties with STAGATE \citep{stagate} at the laminar level (Table~\ref{tab:multiscale}). Existing methods mostly capture part of the hierarchy: STAGATE and Novae score well on region and band but poorly on cell type, whereas BANKSY \citep{banksy} and CellPLM show the opposite pattern. These results show that a single pre-trained backbone can support representations from local cell identity to higher-order tissue organization by simply changing the spatial scale of the readout.

\subsection{Spatial generation of missing cellular configurations}
\label{sec:generation}

We next evaluate the generative capability of \textsc{Sudoku} by asking it to recover missing cell states from partially observed neighborhoods. From 31 human and mouse, brain and kidney sections, we construct 972 questions, each masking part of a 500-cell window in one of four missing geometries (scattered points, interior block, shell, multiple blocks) at one of three missing fractions (0.2, 0.4, 0.6). Questions are further split into easy, medium and hard tertiles by the cell-type composition complexity of the window. Generated cells are mapped to cell types through a code-to-type table built outside the window, and the metric is the cell-type accuracy of the generated cells. Besides GeST, the only existing spatial generative model, we compare with baselines that assign each masked cell the majority type of its $k$ nearest visible cells ($k$-NN, $k=1,5$). Details on benchmark construction, code-to-type mapping and intuitive visualization of the questions are in Appendix~\ref{app:benchmarks}.

Though accuracy falls with composition difficulty for every method, \textsc{Sudoku} achieves the highest accuracy in every geometry-by-difficulty setting and every tissue group (Table~\ref{tab:generation}). GeST stays below the $k$-NN baselines in many settings. These results show that \textsc{Sudoku} recovers missing cellular configurations across organs, species and geometries. We report more baselines and the standard deviations over three seeds in Appendix~\ref{app:generation-extra}.

\begin{table}[h]
\centering
\caption{Spatial generation, cell-type accuracy of the generated cells on the 972 questions, split by missing geometry and difficulty (left) and by tissue group (right). AD: Alzheimer's disease; Ctrl: control; M1, M2: two mice; Inj.: injured. \textbf{Bold}: best; \underline{underline}: second best.}
\label{tab:generation}
\scriptsize
\setlength{\tabcolsep}{1.3pt}
\begin{tabular*}{\linewidth}{@{\extracolsep{\fill}} l ccc ccc ccc ccc cc cc cc @{}}
\toprule
& \multicolumn{12}{c}{By missing geometry and difficulty} & \multicolumn{6}{c}{By tissue group} \\
\cmidrule(lr){2-13}\cmidrule(lr){14-19}
& \multicolumn{3}{c}{Scattered} & \multicolumn{3}{c}{Interior block} & \multicolumn{3}{c}{Shell} & \multicolumn{3}{c}{Multiple blocks} & \multicolumn{2}{c}{Human brain} & \multicolumn{2}{c}{Mouse brain} & \multicolumn{2}{c}{Mouse kidney} \\
\cmidrule(lr){2-4}\cmidrule(lr){5-7}\cmidrule(lr){8-10}\cmidrule(lr){11-13}\cmidrule(lr){14-15}\cmidrule(lr){16-17}\cmidrule(lr){18-19}
Method          & Easy              & Med.              & Hard              & Easy              & Med.              & Hard              & Easy              & Med.              & Hard              & Easy              & Med.              & Hard              & AD                & Ctrl              & M1                & M2                & Inj.              & Ctrl              \\
\midrule
1-NN            & 0.450             & 0.339             & 0.215             & 0.433             & 0.307             & 0.198             & 0.434             & 0.323             & 0.202             & 0.448             & 0.315             & 0.199             & 0.439             & 0.372             & 0.249             & 0.297             & 0.269             & 0.382             \\
5-NN            & 0.474             & \underline{0.346} & \underline{0.251} & \underline{0.502} & \underline{0.368} & 0.232             & \underline{0.488} & \underline{0.324} & \underline{0.253} & \underline{0.504} & \underline{0.330} & 0.246             & \underline{0.514} & \underline{0.457} & \underline{0.266} & \underline{0.314} & 0.296             & 0.377             \\
\midrule
GeST            & \underline{0.503} & 0.321             & 0.249             & 0.476             & 0.326             & \underline{0.261} & 0.486             & 0.282             & 0.249             & 0.483             & 0.296             & \underline{0.252} & 0.485             & 0.444             & 0.228             & 0.294             & \underline{0.317} & \underline{0.396} \\
\midrule
\textsc{Sudoku} & \textbf{0.508}    & \textbf{0.378}    & \textbf{0.267}    & \textbf{0.534}    & \textbf{0.390}    & \textbf{0.263}    & \textbf{0.498}    & \textbf{0.376}    & \textbf{0.282}    & \textbf{0.521}    & \textbf{0.373}    & \textbf{0.278}    & \textbf{0.563}    & \textbf{0.469}    & \textbf{0.290}    & \textbf{0.344}    & \textbf{0.320}    & \textbf{0.406}    \\
\bottomrule
\end{tabular*}
\end{table}

\subsection{Predicting immune recruitment after spatial perturbation}
\label{sec:spac}

Finally, we ask whether the learned conditional distribution predicts the immune recruitment around perturbed tumor clones from the state of the clone alone. We focus on immune recruitment around perturbed tumor clones: whether a tumor draws immune cells into its surroundings separates hot tumors from cold ones and shapes the response to immunotherapy \citep{galon2019,tumeh2014}. SPAC-seq profiles mouse lung metastases whose tumor clones each carry a CRISPR perturbation. For each of 125 clones, we mask cells around it and ask the model to generate them, without telling the model which perturbation the clone carries. The immune-cell fraction among the generated cells is the predicted recruitment of the clone. We report the Spearman correlation between predicted and measured recruitment across the clones of a section, one value for each of the three sections, and the rank error of each clone, the absolute difference between its predicted and measured rank divided by the number of clones in the section, one value per clone. GeST and a simple expression-interpolation baseline are used for comparison. Details of benchmark construction, baseline settings and metrics are in Appendix~\ref{app:benchmarks}.

\begin{figure}[t]
\centering
\includegraphics[width=\linewidth]{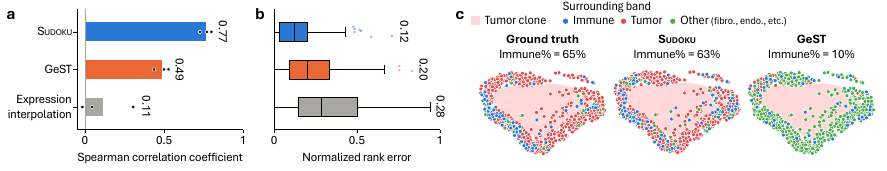}
\caption{Immune recruitment in SPAC-seq. (a) Spearman correlation between predicted and measured recruitment, mean over the three sections, one dot per section. (b) Rank error of each clone; boxes span the quartiles, whiskers 1.5 times the interquartile range, medians written beside the boxes; dots beyond the whiskers are outlier clones, whose rank error exceeds the upper quartile by more than 1.5 times the interquartile range. (c) A Cd274 (PD-L1) knockout clone (pink) and its 50-$\mu$m band, colored by measured type and by the types generated by \textsc{Sudoku} and GeST, with the immune fraction of the band above each.}
\label{fig:spac}
\end{figure}

\begin{wraptable}[18]{r}{0.46\linewidth}
\setlength{\abovecaptionskip}{2pt}\setlength{\belowcaptionskip}{2pt}
\centering
\caption{Ablations on niche retrieval (HESTA AUPRC, mean over all 84{,}977 scoreable pairs) and spatial generation (type accuracy, mean over all 972 questions); each variant changes only the item named.}
\label{tab:ablation}
\scriptsize
\renewcommand{\arraystretch}{0.85}
\setlength{\tabcolsep}{3pt}
\begin{tabular}{l cc}
\toprule
Variant                         & Representation & Generation     \\
\midrule
\multicolumn{3}{l}{w/o residual quantization}                     \\
\quad one level, 256 codes      & 0.467          & 0.305          \\
\quad one level, 65{,}536 codes & 0.441          & 0.247          \\
\midrule
\multicolumn{3}{l}{w/o generative objective}                      \\
\quad latent regression         & 0.429          & 0.368          \\
\quad latent regression (Gaussian) & 0.378 & 0.370 \\
\quad fixed masking rate 0.15   & 0.489          & 0.379          \\
\midrule
\multicolumn{3}{l}{w/o continuous visible input}                  \\
\quad visible cells as codes    & 0.401          & 0.323          \\
\midrule
\multicolumn{3}{l}{w/o positional encoding}                    \\
\quad RoPE removed           & 0.406   & 0.317   \\
\midrule
\textsc{Sudoku}                 & \textbf{0.503} & \textbf{0.405} \\
\bottomrule
\end{tabular}
\end{wraptable}

\textsc{Sudoku} reaches a Spearman correlation of 0.77 averaged over the three sections, compared with 0.49 for GeST and 0.11 for expression interpolation (Figure~\ref{fig:spac}a). It also ranks the clones by immune recruitment more accurately, with a median rank error of 0.12, compared with 0.20 for GeST and 0.28 for interpolation (Figure~\ref{fig:spac}b). These results show that \textsc{Sudoku} recovers immune recruitment variation across perturbed clones from the clone state. We compare with a direct predictor of immune recruitment and score three spatial readouts of the generated cells in Appendix~\ref{app:spac-extra}. As a case study, Figure~\ref{fig:spac}c shows a Cd274 knockout clone. Cd274 encodes PD-L1, through which tumor cells switch off PD-1-bearing T cells, and the PD-1/PD-L1 axis is the target of checkpoint blockade immunotherapy \citep{tumeh2014,galon2019}, so a clone that has lost it is expected to be heavily infiltrated. The measured immune fraction in its band is 0.65. \textsc{Sudoku} generates 0.63 with a spatial pattern close to the measured one, whereas GeST generates 0.10 and collapses much of the band onto one non-immune type.

\subsection{Ablations}
\label{sec:ablations}

We ablate \textsc{Sudoku} along four axes (Table~\ref{tab:ablation}), all variants sharing the backbone, corpus and schedule: the two-level vocabulary is replaced by one level with 256 or 65{,}536 codes; the generative objective is replaced by two predictive controls, regressing the latent of a masked cell with a squared-error or a Gaussian head instead of predicting a distribution over its codes (Appendix~\ref{app:ablation}), or fixing the masking rate at 0.15; visible cells enter as their codes instead of their continuous latents; and the positional encoding is removed.

Residual quantization matters most for generation, where one level of either size loses up to 0.16 accuracy. Regressing the latent instead of predicting a distribution hurts both representation and generation, and a fixed masking rate mainly hurts generation. Feeding visible cells as codes costs 0.10 niche retrieval AUPRC and 0.08 cell type generation accuracy. Relative to the variant without positional encoding, \textsc{Sudoku} improves niche retrieval AUPRC by 23.9\% (0.406 to 0.503) and generation accuracy by 12.1\% (0.372 to 0.417) when both take the most probable cell type (Appendix~\ref{app:generation-extra}), indicating that \textsc{Sudoku} uses where each cell is, beyond the cell composition of its neighborhood. Part of the larger generation gap in Table~\ref{tab:ablation} (0.317 to 0.405) comes from reading types through the most probable code. The hierarchical vocabulary, the generative objective over the whole masking-rate axis, the continuous visible input and the positional encoding thus each carry part of the performance of \textsc{Sudoku}.
\section{Discussion}
\label{sec:conclusion}

We present \textsc{Sudoku}, a spatial foundation model that learns tissue organization through spatial discrete denoising. Beyond a new pre-training objective, the results support the view of the introduction: tissue organization can be learned as conditional cellular distributions under arbitrary partial observation. Varying how much of a neighborhood is hidden exposes a continuum of spatial dependencies, from local cellular constraints to collective multicellular organization, and representation and generation arise as two regimes of this one conditional model rather than as separate objectives. On representation, generation and perturbation-response benchmarks, \textsc{Sudoku} outperforms existing pre-trained models, specialized methods and autoregressive generation.

\textsc{Sudoku} still has several limitations. The tokenizer is an MLP over the whole transcriptome and therefore requires an input of fixed dimension. A design that accepts partial gene panels should be explored. Applying \textsc{Sudoku} to new sections still requires self-supervised fine-tuning, and a larger pre-training corpus covering more platforms and spatial resolutions may improve its zero-shot generalization. The spatial generation benchmark scores each generated cell by its type and does not yet measure neighborhood-level structure. Future evaluation will compare multi-scale neighborhood composition and cell-type co-occurrence between generated and measured tissue, in disease-associated neighborhoods and under in silico spatial perturbation. Finally, \textsc{Sudoku} learns tissue organization from static spatial transcriptomic profiles. Extending it to tissue dynamics and to morphology, histology and other modalities, toward a model of tissue as a whole, is an important future direction.

\bibliography{refs}
\bibliographystyle{iclr2027_conference}

\clearpage
\appendix
\setcounter{figure}{0}\renewcommand{\thefigure}{A.\arabic{figure}}
\setcounter{table}{0}\renewcommand{\thetable}{A.\arabic{table}}
\section{Pre-training details}
\label{app:tokenizer}

\paragraph{Pre-training corpus.} The corpus is 396 cell-segmented whole-transcriptome Stereo-seq sections, 161 human and 235 mouse, covering 12 organs (brain, skin, intestine, eye, heart, liver, stomach, breast, nasopharynx, lung, bone and ovary), 356 normal and 40 tumor, 49.1 million cells. This corpus is what remains of about 80 million cells after the following quality control: a cell with fewer than 100 detected genes is dropped, and a section is dropped when this removes more than 20\% of its cells. Genes are restricted to the 15{,}903 protein-coding genes with a one-to-one human--mouse ortholog, so one vocabulary serves both species. A cell is a neighborhood center in pre-training only when at least 10 cells lie within 150\,$\mu$m of it.

\paragraph{Hyperparameters.} Table~\ref{tab:hyper} lists the tokenizer, the backbone and the pre-training schedule. The tokenizer uses cosine distance at the first level and Euclidean distance on the residual at the second, and each of its batches draws one cell from each of 1{,}024 sections with replacement, so that no section dominates the codebook statistics of a batch. The tokenizer trains for 1.8 hours and the backbone for 3.4 hours, each on one A100 80G GPU.

\paragraph{Second-level head.} The two classification heads of Section~\ref{sec:pretraining} predict the two code levels independently, in one forward pass with one head per level. MusicGen \citep{musicgen} uses a similar design for residual-quantized audio tokens. A variant whose second-level head also takes the first-level code as input, trained with teacher forcing, gives little gain over the independent heads (Table~\ref{tab:cond-head}), so \textsc{Sudoku} keeps the simpler design.

\begin{table}[htb]
\centering
\caption{Hyperparameters of the tokenizer, the backbone, pre-training and decoding.}
\label{tab:hyper}
\footnotesize
\setlength{\tabcolsep}{6pt}
\begin{tabular}{l p{3.25in}}
\toprule
\multicolumn{2}{l}{\textit{Tokenizer}} \\
Encoder / decoder widths           & 512, 256, 128 / 256, 512                                      \\
Latent dimension                   & 128                                                           \\
Codebooks                          & 2 levels $\times$ 256 codes                                   \\
Distance per level                 & cosine, Euclidean                                             \\
Codebook update                    & exponential moving average, decay 0.99                        \\
Commitment weight $\beta$          & 1.0                                                           \\
Normalization                      & BatchNorm                                                     \\
Batch                              & 1{,}024 sections $\times$ 1 cell                              \\
Steps, learning rate               & 90{,}000, $10^{-3}$                                           \\
Weight decay, gradient clipping    & 0.01, 1.0                                                     \\
\midrule
\multicolumn{2}{l}{\textit{Backbone}} \\
Layers, width, heads, MLP width    & 10, 256, 8, 1{,}024                                           \\
Positional encoding                & axial RoPE, $\theta=10$                                       \\
Neighborhood radius, minimum cells & 150\,$\mu$m, 10                                               \\
\midrule
\multicolumn{2}{l}{\textit{Pre-training}} \\
Masking rate                       & $t\sim U(0.001, 1)$                                           \\
Batch                              & 768 neighborhoods $\times$ 2 accumulation steps               \\
Steps, learning rate               & 32{,}500, $10^{-4}$ constant                                  \\
Optimizer                          & $\beta=(0.9, 0.95)$, weight decay 0.01, gradient clipping 1.0 \\
\midrule
\multicolumn{2}{l}{\textit{Decoding}} \\
Steps $T$, temperature $\tau$      & 32, 0 \\
\bottomrule
\end{tabular}
\end{table}

\begin{table}[htb]
\centering
\caption{Second-level head, with everything else fixed; metrics as in Table~\ref{tab:ablation}.}
\label{tab:cond-head}
\footnotesize
\begin{tabular}{l cc}
\toprule
Second-level head                                  & Representation & Generation \\
\midrule
Independent (\textsc{Sudoku})                      & 0.503          & 0.405      \\
Conditioned on the first-level code                & 0.501          & 0.410      \\
\bottomrule
\end{tabular}
\end{table}

\clearpage
\section{\texorpdfstring{Codebook usage}{Codebook usage}}
\label{app:codebook}

\begin{figure}[htb]
\centering
\includegraphics[width=\linewidth]{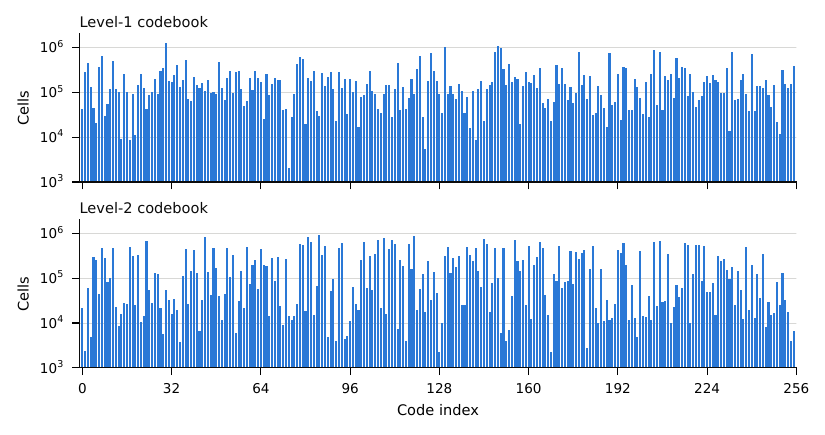}
\caption{Number of pre-training cells on each code of the level-1 and level-2 codebooks, on a logarithmic axis.}
\label{fig:codebook-usage}
\end{figure}

We count the codes of all 49.1 million cells of the pre-training corpus to check that the codebooks are not collapsed. Every code at both levels is in use, the least used one by 2{,}075 cells at level 1 and 2{,}263 cells at level 2 (Figure~\ref{fig:codebook-usage}), and 39{,}718 of the 65{,}536 two-level code pairs are used (Table~\ref{tab:codebook-usage}).

We measure the diversity of the codes by the cosine similarity of every pair of codes in a codebook (Table~\ref{tab:codebook-similarity}). The most similar pair has a cosine similarity of 0.794 at level 1 and 0.809 at level 2, so no two codes are near duplicates. The median pairwise cosine similarity is close to zero at both levels, so the codes point in diverse directions instead of crowding together.

\begin{table}[htb]
\centering
\begin{minipage}[t]{0.47\linewidth}
\centering
\caption{Codebook utilization over the 49.1 million cells of the pre-training corpus.}
\label{tab:codebook-usage}
\footnotesize
\begin{tabular}{l rrr}
\toprule
              & Used   & Total  & Utilization \\
\midrule
Level-1 codes & 256    & 256    & 100.0\%     \\
Level-2 codes & 256    & 256    & 100.0\%     \\
Code pairs    & 39{,}718 & 65{,}536 & 60.6\%  \\
\bottomrule
\end{tabular}
\end{minipage}\hfill
\begin{minipage}[t]{0.47\linewidth}
\centering
\caption{Cosine similarity among the 32{,}640 pairs of codes in each codebook.}
\label{tab:codebook-similarity}
\footnotesize
\begin{tabular}{l cc}
\toprule
                  & Level 1 & Level 2 \\
\midrule
Median cosine similarity  & 0.019   & 0.003   \\
Largest cosine similarity & 0.794   & 0.809   \\
\bottomrule
\end{tabular}
\end{minipage}
\end{table}

\clearpage
\section{Rotation of the coordinates at test time}
\label{app:rotation}

Theoretically, the 2D RoPE of Section~\ref{sec:pretraining} is not rotation-invariant, since the two halves of each head are rotated by the $x$ and $y$ components of the displacement separately. In practice, sections are placed on the slide at arbitrary orientations, so we test how much rotating the coordinates at test time affects the model performance. We keep the fine-tuned \textsc{Sudoku} weights and rotate only the center-relative coordinates that enter the positional encoding, at degrees including 45, 90, 158, 180 and 279 (158 and 279 degrees are drawn at random). The neighborhood graphs, the measured spacing and the tokenizer codes remain unchanged. The 0-degree setting thus refers to the original model. Table~\ref{tab:rotation} reports the results. The representation moves little at every stage of the readout. The cosine similarity between the rotated and the unrotated run is near 1 in all settings. As a comparison, we also randomly draw representations of different items without rotation, whose median cosine similarity is only 0.19, 0.30 and 0.07. In downstream performance metrics, HESTA AUPRC almost has no change as well. For the generation accuracy, it moves by at most 0.008. These results show that in practice, the model is robust to arbitrary rotations of the coordinates.

\begin{table}[htb]
\centering
\caption{Impact of test-time coordinate rotation on the model. Top: median cosine similarity between the rotated and the unrotated run of the same item at each stage of the HESTA readout; the last column is the median cosine between two different items at 0 degrees. Bottom: benchmark scores at each angle. --: not exported.}
\label{tab:rotation}
\footnotesize
\setlength{\tabcolsep}{3pt}
\begin{tabular}{l cccccc c}
\toprule
& \multicolumn{6}{c}{Rotation angle (degrees)} & Different \\
\cmidrule(lr){2-7}
                                               & 0     & 45      & 90      & 158     & 180     & 279     & items \\
\midrule
\multicolumn{8}{l}{\emph{Cosine similarity to the unrotated run, median}} \\
Cell representation $\mathbf{h}$               & 1     & 0.9976  & 0.9953  & --      & 0.9941  & --      & 0.19  \\
Neighborhood representation $\bar{\mathbf{h}}$ & 1     & 0.9999  & 0.9998  & 0.9997  & 0.9997  & 0.9998  & 0.30  \\
Niche query vector                             & 1     & 0.99999 & 0.99999 & 0.99998 & 0.99998 & 0.99999 & 0.07  \\
\midrule
\multicolumn{8}{l}{\emph{Benchmark readings}} \\
HESTA AUPRC                                    & 0.503 & 0.503   & 0.503   & 0.502   & 0.502   & 0.503   &       \\
Generation accuracy                            & 0.405 & 0.401   & 0.397   & 0.397   & 0.397   & 0.399   &       \\
\bottomrule
\end{tabular}
\end{table}

\clearpage
\section{Training objective as a variational bound}
\label{app:bound}

Section~\ref{sec:pretraining} feeds visible cells as continuous latents but predicts discrete codes, which differs from masked diffusion language models \citep{mdlm,radd}, where both the input and the target are codes. This appendix shows that Equation~\ref{eq:loss} is nevertheless the variational bound of an absorbing-state diffusion model over the neighborhood, so minimizing it trains a generative model. The argument has three steps: the token is defined as the pair of a latent and its code, so that a diffusion over tokens may condition on latents; the reverse process is factorized so that the network only predicts the code; and the extra term this introduces into the bound does not depend on the network.

\paragraph{State space.} Fix the neighborhood $N$ and its coordinates $\mathbf{p}_N$; only cell states diffuse, and every distribution below is conditioned on $\mathbf{p}_N$. The token of cell $j$ is the pair $\mathbf{x}_j=(\mathbf{z}_j,c_j)$ of its tokenizer latent and its two-level code $c_j=Q(\mathbf{z}_j)$, a deterministic function of the latent under the frozen tokenizer. The data distribution is the distribution $q(\mathbf{x}_N)$ of these pairs over corpus neighborhoods.

\paragraph{Forward and reverse process.} The absorbing forward process replaces each token independently by the mask with probability $t$; it acts on whole tokens and does not depend on what a token contains. Writing $M_t$ for the masked set and $\mathbf{x}_t$ for the partially masked neighborhood, the reverse process of \citet{mdlm} unmasks each masked cell independently and draws its token from a model $p_\theta(\mathbf{x}_j\mid\mathbf{x}_t)$. We take
\begin{equation}
p_\theta(\mathbf{x}_j\mid\mathbf{x}_t)=p_\theta(c_j\mid\mathbf{x}_t)\;q_{\mathrm{tok}}(\mathbf{z}_j\mid c_j),
\label{eq:reverse-factor}
\end{equation}
where $q_{\mathrm{tok}}(\mathbf{z}\mid c)$ is the distribution of tokenizer latents within the cell of code $c$ over the corpus, a fixed distribution that does not depend on $\theta$. Since $c_V=Q(\mathbf{z}_V)$, the partially masked neighborhood $\mathbf{x}_t$ carries exactly the information of $(\mathbf{z}_V,\mathbf{p}_N)$, so $p_\theta(c_j\mid\mathbf{x}_t)$ is the network of Section~\ref{sec:pretraining}, read at the masked cells.

\paragraph{The bound.} For the absorbing process with the linear schedule, the negative evidence lower bound of \citet{mdlm} and \citet{radd} in continuous time is
\begin{equation}
-\log p_\theta(\mathbf{x}_N)\;\le\;\mathbb{E}_{t\sim U(0,1)}\,\mathbb{E}_{\mathbf{x}_t}\;\frac{1}{t}\sum_{j\in M_t}-\log p_\theta(\mathbf{x}_j\mid\mathbf{x}_t).
\label{eq:nelbo}
\end{equation}
Substituting Equation~\ref{eq:reverse-factor} splits the summand into $-\log p_\theta(c_j\mid\mathbf{z}_V,\mathbf{p}_N)-\log q_{\mathrm{tok}}(\mathbf{z}_j\mid c_j)$. The second term does not depend on $\theta$, so the right-hand side equals Equation~\ref{eq:loss} plus a constant, and minimizing Equation~\ref{eq:loss} minimizes the bound. The $1/t$ weighting is the schedule term of the bound. Equation~\ref{eq:loss} is therefore the negative evidence lower bound of a generative model over neighborhoods, up to a constant, which is the property used in Section~\ref{sec:pretraining}.

\paragraph{Relation to the sampler.} The bound above concerns the model with the reverse process of Equation~\ref{eq:reverse-factor}, in which the unmasked cells are chosen at random and the latent of an unmasked cell is drawn from $q_{\mathrm{tok}}(\mathbf{z}\mid c)$. The sampler of Section~\ref{sec:inference} uses the same network but approximates this reverse process in three ways: a committed cell enters the conditioning through its codebook reconstruction $\mathbf{z}_q(\hat{c}_j)$ rather than through a draw from $q_{\mathrm{tok}}(\mathbf{z}\mid c)$, cells are committed in order of confidence rather than at random, and the cells committed within one step, as well as the two code levels of a cell, are drawn independently given the current context. Pre-training also draws $t$ from $U(\epsilon,1)$ with $\epsilon=0.001$ rather than from $U(0,1)$.

\clearpage
\section{Benchmarks}
\label{app:benchmarks}

Table~\ref{tab:benchmarks} lists the data of the five benchmarks.

\begin{table}[htb]
\centering
\caption{The evaluation suite. Protocols are in Appendix~\ref{app:benchmarks} and Appendix~\ref{app:baselines}, SPAC-seq in three folds by section, and no benchmark section is in the pre-training corpus.}
\label{tab:benchmarks}
\scriptsize
\setlength{\tabcolsep}{3pt}
\begin{tabular}{@{}l l l l l r r l@{}}
\toprule
Benchmark              & Species      & Tissue          & Platform   & Resolution & Sections & Cells        & Units          \\
\midrule
HESTA                  & human        & embryo          & Stereo-seq & cell       & 23       & 9.7M         & 5,650 NOIs     \\
DLPFC                  & human        & cortex          & 10x Visium & spot       & 12       & 47,329 spots & 252 NOIs       \\
Multi-scale clustering & mouse        & brain           & MERFISH    & cell       & 6        & 235,857      & 3 label levels \\
Spatial generation     & human, mouse & brain, kidney   & Stereo-seq & cell       & 31       & 1,938,543    & 972 questions  \\
SPAC-seq               & mouse        & lung metastasis & Visium HD & cell       & 3        & 597,007      & 125 clones     \\
\bottomrule
\end{tabular}
\end{table}

\paragraph{Protocols.} No benchmark section is in the pre-training corpus. \textsc{Sudoku} is fine-tuned on the benchmark sections with its own objective and no benchmark label; how each baseline is run is stated in Appendix~\ref{app:baselines}. Benchmark sections enter \textsc{Sudoku} through the pre-training vocabulary of 15{,}903 genes: genes are matched by identifier or through the one-to-one ortholog table, and missing genes are set to zero. A Visium spot of DLPFC is treated as one cell. On HESTA, DLPFC and multi-scale clustering the sections fine-tuned on are the sections scored, all 23 of HESTA, all 12 of DLPFC and the six MERFISH sections. On spatial generation the sections fine-tuned on are 61 sections of the same three cohorts with no questions. On SPAC-seq each fold fine-tunes the model on its two training sections, builds its code-to-type table there, and generates on the held-out section. Fine-tuning has two steps (Table~\ref{tab:finetune}). On every benchmark except spatial generation, the tokenizer is first fine-tuned on the benchmark sections with its own objective, with all its parameters and codebooks updated. All parameters of the spatial transformer are then fine-tuned with a learning rate of $5\times10^{-5}$ and the same masking-rate distribution as pre-training, $t\sim U(0.001,1)$, with the codebooks fixed.

\begin{table}[htb]
\centering
\caption{Fine-tuning protocol of \textsc{Sudoku} per benchmark. Step counts and times are rounded.}
\label{tab:finetune}
\footnotesize
\setlength{\tabcolsep}{4pt}
\begin{tabular}{l l r r}
\toprule
Benchmark              & Tokenizer                                 & Spatial transformer & Compute on one A100 \\
\midrule
HESTA                  & 1{,}000 steps, learning rate $5\times10^{-5}$ & 38{,}000 steps      & 1.5\,h              \\
DLPFC                  & 3{,}000 steps, learning rate $5\times10^{-5}$ & 2{,}000 steps       & 10\,min             \\
Multi-scale clustering & 1{,}000 steps, learning rate $5\times10^{-5}$ & 10{,}000 steps      & 20\,min             \\
Spatial generation     & unchanged                                 & 3{,}000 steps       & 15\,min             \\
Immune recruitment     & 1{,}000 steps, learning rate $5\times10^{-5}$ & 75{,}000 steps      & 4\,h                \\
\bottomrule
\end{tabular}
\end{table}

\begin{figure}[t]
\centering
\includegraphics[width=\linewidth]{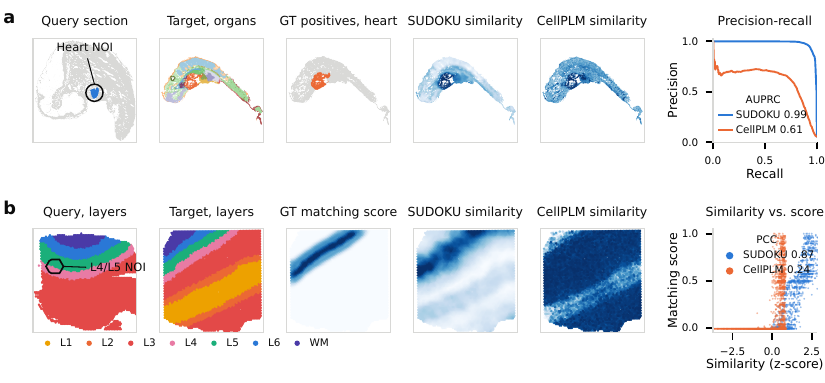}
\caption{Niche retrieval, one worked example per scenario, \textsc{Sudoku} next to CellPLM. (a) HESTA. A heart niche of interest (NOI) on one section is the query, and the target section is shown colored by organ. Every candidate niche of the target is scored by the cosine similarity of its representation to the query NOI representation, shown as a map per model. Candidate niches of the correct organ are the positives, and others are the negatives for the binary classification. (b) DLPFC. An NOI at the L4/L5 junction is the query, and the target section is shown with its cortex layer labels. Every candidate niche of the target is scored by cosine similarity to the query NOI, per model. The ground truth is the matching score between the NOI and each candidate niche, computed by a Weisfeiler-Lehman graph kernel on the layer labels. The score of the pair is the Pearson correlation between similarity and matching score over the candidate niches of the target. Similarity maps run from light to dark over the 1st to 99th percentile of each model's values.}
\label{fig:retrieval-example}
\end{figure}

\paragraph{Niche retrieval, HESTA.} HESTA is 23 Stereo-seq sections of human embryos at cell resolution, 9.7 million cells with an organ label each, from the atlas of \citet{hesta}. We define NOIs in spatially contiguous regions where all cells are purely labeled as the same organ. On each section and for each organ we take the cells of that organ, keep the largest connected component of their 8-nearest-neighbor spatial graph, and split it by k-means on the coordinates into pieces of about 1{,}000 cells, keeping the pieces of 300 to 3{,}000 cells. This gives 5{,}650 niches of interest over 47 organs. The readout radius is the neighborhood radius of pre-training, 150\,$\mu$m (Table~\ref{tab:hyper}). An NOI of about 1{,}000 cells is about 0.5\,mm across, a size that balances two needs of the benchmark. HESTA is annotated at the organ level, so an NOI should be a piece of tissue large enough to reflect the organization of an organ region rather than a few cells. It should also fit inside one organ, so that small organs still give several NOIs per section; at this size, ganglion and heart give about four NOIs per section. This scale is also close to that of the DLPFC NOIs released with QueST, 100 spots that each hold several cells. We show in Appendix~\ref{app:retrieval-extra} that our conclusion does not depend on the NOI size. For an NOI and a target section, every candidate niche on the target sections is scored by the cosine similarity of its representation to the query NOI representation. The cosine similarity then serves as a binary classification metric to identify regions with the correct organ label. The score of the pair (NOI, target section) is the AUPRC score of this binary classification problem. Figure~\ref{fig:retrieval-example}a walks through one pair for \textsc{Sudoku} and CellPLM, a heart NOI queried against a section of an earlier stage. \textsc{Sudoku}'s similarity is high on the heart of the target and low elsewhere, and CellPLM's is high over most of the target. Among the top-ranked target cells, as many as there are heart cells, 94\% are heart for \textsc{Sudoku} and 67\% for CellPLM, and the AUPRC of the pair is 0.99 and 0.61. Pairs whose target section has no cell of the query organ are dropped, leaving 84{,}977 scoreable pairs. Table~\ref{tab:retrieval} reports the mean AUPRC of the eight main organs, which hold 86\% of the scoreable pairs.

\paragraph{Niche retrieval, DLPFC.} DLPFC is twelve 10x Visium sections of human dorsolateral prefrontal cortex from three donors with a cortical layer label per spot, 47{,}329 labeled spots in all \citep{dlpfc}. We use the niche query benchmark released with QueST \citep{quest}, whose NOIs are sets of 100 spots at the junction of two or three layers, seven classes in all (L3L4, L4L5, L5L6, L6WM, L3L4L5, L4L5L6, L5L6WM), with three replicates per class on every section. We also follow QueST for the readout radius, 318\,$\mu$m, which covers about 37 spots. For an NOI and a target section, every candidate niche on the target section is scored by the cosine similarity of its representation to the query NOI representation. The ground truth is the matching score between the NOI and each candidate niche, computed by a Weisfeiler-Lehman graph kernel on the layer labels and released with the benchmark. The score of the pair (NOI, target section) is the Pearson correlation between similarity and matching score over the candidate niches of the target section. Figure~\ref{fig:retrieval-example}b walks through one pair for \textsc{Sudoku} and CellPLM. The matching score on the target section is a band along the border of L4 and L5, with a mean of 0.62 in L4, 0.67 in L5, 0.13 in L6 and near zero in the other layers. \textsc{Sudoku}'s similarity forms the same band, with layer means of 0.32 in L4, 0.42 in L5, 0.13 in L6 and below zero elsewhere, PCC 0.87. CellPLM's similarity varies by less than 0.08 across the layers, PCC 0.24. Every NOI is paired with the other eleven sections, giving 2{,}772 pairs, and Table~\ref{tab:retrieval} reports the mean PCC over the pairs of each class.

\begin{figure}[t]
\centering
\includegraphics[width=\linewidth]{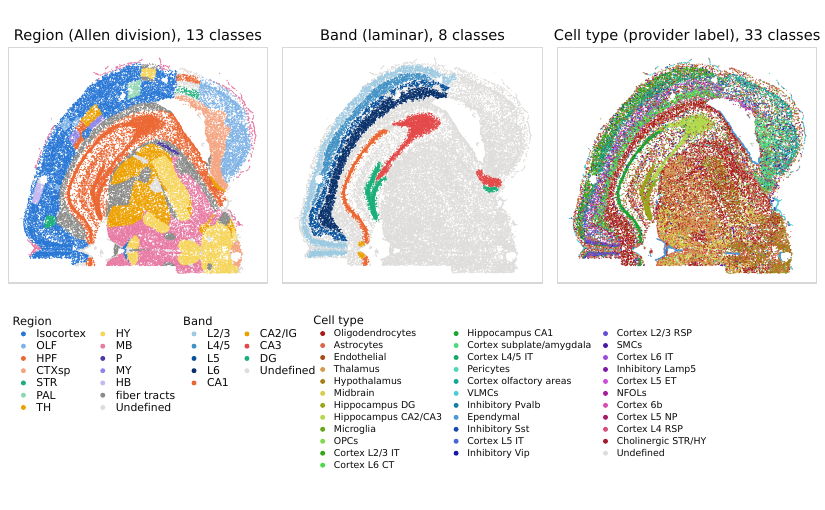}
\caption{The three label levels of the multi-scale clustering benchmark on one section. Region is the Allen brain division of the registered position, Band the laminar band read from the layer-specific neurons around each cell, and Cell type the clustering label of the data provider. Cells without a label at a level are grey.}
\label{fig:multiscale-labels}
\end{figure}

\paragraph{Multi-scale clustering.} The data are six coronal sections of mouse brain from three mice measured by whole-transcriptome MERFISH with 22{,}930 genes and 235{,}857 cells \citep{wtmerfish}. Every cell carries labels at three levels. Region is the Allen brain division of the cell's registered position (isocortex, hippocampal formation, olfactory areas, striatum, pallidum, thalamus, hypothalamus, midbrain, pons, medulla and others), with all fiber tracts merged into one class and all ventricles into another, 12 to 13 classes per section. Band is a laminar band read from the tissue itself. The whole transcriptome of every cell is mapped to the Allen whole-mouse-brain taxonomy \citep{allenabc}, and the neurons whose subclass belongs to one cortical layer (L2/3, L4/5, L5, L6, L6b) or one hippocampal subfield (CA1, CA2 and IG, CA3, DG) vote for the band of every cell within 75\,$\mu$m of them. A cell takes the winning band when at least five such neurons are within reach and the winner holds at least 60\% of the votes, so glia and vessels carry the band of the layer they sit in, 6 to 8 classes per section. Cell type is the clustering label of the data provider, 30 to 36 classes per section. Figure~\ref{fig:multiscale-labels} shows the three levels on one section. At every level a class with fewer than 20 cells or 0.1\% of the cells of a section is undefined there, and undefined cells take part in clustering but not in scoring. A method returns one vector per cell and scale. For each section and each representation we build a 15-nearest-neighbor graph in the representation space and cluster it with Leiden, searching the resolution by bisection so that the number of clusters among the defined cells equals the number of classes of the level being scored. ARI and NMI are computed on the defined cells and Table~\ref{tab:multiscale} reports the median over the six sections.

\begin{figure}[p]
\centering
\includegraphics[width=\linewidth]{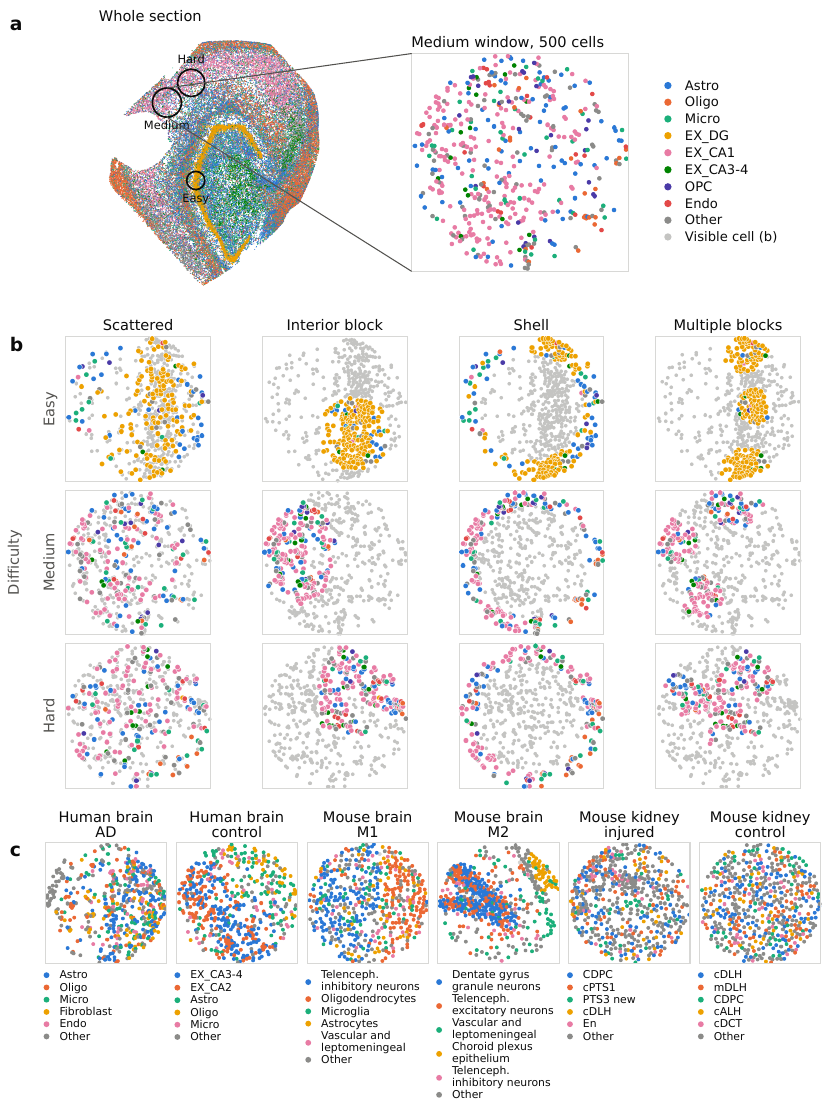}
\caption{Spatial generation questions. (a) A human control brain section with an easy, a medium and a hard window of 500 cells circled, the medium one magnified, all cells colored by true type. (b) The four missing geometries (columns, in the order of Table~\ref{tab:generation}) at missing fraction 0.4 on the three windows (rows). Grey dots are the visible cells given to the model, colored dots the cells to be generated, colored by their true type. (c) One window per tissue group with all cells colored by type. The windows in (b) are the most type-diverse of their tier on one section, and those in (c) the most type-diverse of their group.}
\label{fig:generation-questions}
\end{figure}

\paragraph{Spatial generation.} The benchmark has 31 sections in six groups: human Alzheimer brain (7), human control brain (7), two mouse brains (5 each), injured kidney (4), control kidney (3). Windows of 500 cells are cut per group, 81 in total; four missing geometries times three missing fractions give 972 questions. Scattered points leave visible cells next to every masked cell, the interior block and the shell leave context on one side of the masked region only, and multiple blocks fall in between. Within each tissue group, windows are ranked by a composite heterogeneity score, mainly the entropy of the cell-type composition and the share of the most frequent type, together with the same statistics on the tokenizer codes, and split into easy, medium and hard tertiles. Figure~\ref{fig:generation-questions} shows the four geometries on an easy, a medium and a hard window at missing fraction 0.4, and one window per tissue group. The ground truth is the annotated type of the masked cells. \textsc{Sudoku} decodes with temperature $\tau=0$ (Table~\ref{tab:hyper}), so every committed code is the most probable one, decoding is greedy and no sampling is involved. GeST is decoded in the same way, by taking its most probable prediction for each masked cell. For \textsc{Sudoku} and GeST, each generated cell is mapped to a single type through a code-to-type table built on the other sections of the same tissue group, so the argmax is taken over codes, not over types; Appendix~\ref{app:generation-extra} also reports the most probable type. All methods are scored in the same way: each masked cell is assigned a single type, and accuracy is the fraction of masked cells whose assigned type matches the annotated one.

\paragraph{Immune-recruitment prediction.} From the 345 single-perturbation clones of three sections we keep those whose perturbation has at least two clones with consistent measured recruitment, a coefficient of variation across these clones of at most 0.35, so that the measured recruitment is reproducible enough to serve as ground truth, whose 50-$\mu$m band holds at least 20 immune cells, so that the immune fraction is estimable, and whose clone and band together hold at most 1{,}000 cells, the neighborhood capacity of the model, giving 125 questions. We mask all cells within a 50-$\mu$m band around the clone, give the model only the cells of the clone, generate the states of the cells in the band, and read out the fraction of immune cells among the generated non-tumor cells. The model input carries no perturbation identity and the prediction depends only on spatial context. The readout goes through the codebook: on the fine-tuning sections we estimate the conditional distribution $P(y\mid c)$ from each two-level code to annotated type; writing $I$ for the set of immune types, $y_{\mathrm{tum}}$ for the tumor type and $B$ for the cells to be generated in the band, and $p_\theta(c_j\mid\text{clone})$ for the distribution of cell $j$ at the step it is committed, the predicted immune fraction of a question is the expectation
\begin{equation}
\hat{f}=\frac{\sum_{j\in B}\sum_{c}p_\theta(c_j=c\mid\text{clone})\,P(y\in I\mid c)}{\sum_{j\in B}\sum_{c}p_\theta(c_j=c\mid\text{clone})\,P(y\neq y_{\mathrm{tum}}\mid c)}.
\end{equation}
The metrics are the Spearman correlation between $\hat{f}$ and the measured fraction over the clones of each section and the rank error of each clone, as defined in Section~\ref{sec:spac}.

\clearpage
\section{Baselines}
\label{app:baselines}

\paragraph{Baselines.} Every external model is run from its pre-trained weights with its default settings and read out at the same neighborhood radius as \textsc{Sudoku}. GeST, CellPLM, Nicheformer and Novae, whose training code is available, are further fine-tuned on the benchmark sections with their own self-supervised objectives and no benchmark label; scGPT-spatial, SToFM and CellWorld release inference code only, and the self-supervised adaptation of TERRA exceeds our compute budget, so these four are used as released. Where a metric needs a type, a model's generated cells are typed through a table built on the same reference cells as our code-to-type table. Human sections are matched to each model's vocabulary by Ensembl identifier or gene symbol, and the mouse sections of the multi-scale benchmark through the one-to-one ortholog table, with genes outside a vocabulary dropped.
\emph{CellPLM} \citep{cellplm} starts from the released 85M checkpoint, is fine-tuned with its own objective on the benchmark sections, and is read out through its 512-dimensional cell embedding.
\emph{Nicheformer} \citep{nicheformer} uses the released weights and is read out as the mean of the last layer over the gene tokens; its output head is not released and is trained for one epoch of its masked gene modeling on the benchmark sections, and HESTA is given a reserved platform token since Stereo-seq is not among its platforms.
\emph{scGPT-spatial} \citep{scgptspatial} and \emph{SToFM} \citep{stofm} use their released weights and cell embeddings; SToFM tiles each section into sub-slices of about 1{,}000 cells as in its own pipeline.
\emph{CellWorld} \citep{cellworld} uses the released base model on the 512 most expressed genes of each cell, with the organ token set to brain for the brain sections and to unknown for HESTA, and is read out at each cell's own row of its context encoder.
\emph{TERRA} \citep{terra} uses the released 112M model on each cell with its ten nearest neighbors and is read out through its neighborhood embedding; on the multi-scale benchmark the cell type level uses its cell embedding.
\emph{Novae} \citep{novae} uses the released human weights on HESTA and the released mouse weights on the multi-scale benchmark, each fine-tuned with its own objective on the benchmark sections.
\emph{GeST} \citep{gest} starts from a checkpoint pre-trained on the same corpus as \textsc{Sudoku} and is fine-tuned with its own objective on each benchmark. Downstream tasks use its default settings.
\emph{QueST} \citep{quest}, \emph{STAGATE} and \emph{BANKSY} \citep{stagate,banksy} are task-specific methods without pre-training, designed for niche retrieval and spatial domain detection on ST data. We use their default settings.
The training-free baselines are $k$-NN on spatial generation, which assigns each masked cell the majority type of its $k$ nearest visible cells, and expression interpolation on SPAC-seq, which interpolates the expression of each cell in the band from the visible cells and types it by the nearest type centroid. Table~\ref{tab:baseline-protocol} summarizes, for every method, its starting weights and its adaptation to the benchmark sections.

\begin{table}[htb]
\centering
\caption{Starting weights and adaptation of every method. All models are read out at the same neighborhood radius as \textsc{Sudoku}.}
\label{tab:baseline-protocol}
\footnotesize
\begin{tabular}{lll}
\toprule
Methods & Starting weights & Adaptation to the benchmark sections \\
\midrule
\textsc{Sudoku}, GeST                  & pre-trained on our corpus & fine-tuned with its own objective \\
CellPLM, Nicheformer, Novae            & released weights          & fine-tuned with its own objective \\
scGPT-spatial, SToFM, CellWorld        & released weights          & none, inference code only \\
TERRA                                  & released weights          & none, beyond our compute budget \\
QueST, STAGATE, BANKSY                 & none, task-specific       & default settings \\
$k$-NN, expression interpolation       & none, training-free       & none \\
\bottomrule
\end{tabular}
\end{table}

\clearpage
\section{\texorpdfstring{Ablation details}{Ablation details}}
\label{app:ablation}

The latent-regression variant of Table~\ref{tab:ablation} differs from \textsc{Sudoku} only in its output head and loss. It replaces the two classification heads with one linear head that predicts, for each masked cell, its continuous, 128-dimensional tokenizer latent $\mathbf{z}$, and its loss is the mean squared error weighted by $1/t$, where $t$ is the masking rate drawn for each neighborhood, as in Equation~\ref{eq:loss}. Since this variant cannot be sampled, for generation it predicts the latents of all masked cells in one forward pass, quantizes each prediction into a code pair, and reads cell types through the same code-to-type table as \textsc{Sudoku}.

The Gaussian-head variant outputs a mean and a per-dimension variance of $\mathbf{z}$ for each masked cell, and its loss is the Gaussian negative log-likelihood, weighted by $1/t$ in the same way. For generation, its predicted mean is quantized in the same way as in the latent-regression variant.

Quantizing the predicted latent may not be fair to the regression variants, so we also read cell types directly from latents with $k$-NN classifiers, with $k$ from 1 to 10{,}000 (Table~\ref{tab:latentknn}). The classifiers are built on the tokenizer latents of the reference cells, the annotated cells on which the code-to-type table is built. Each method gives the classifier one latent per masked cell. For the regression variants, this is the predicted latent. For \textsc{Sudoku}, which predicts codes instead of latents, it is the sum of the codebook vectors of its generated code at the two levels. At every $k$, \textsc{Sudoku} remains the most accurate, and the $k$-NN classifier brings little gain to either regression variant, so the gap between discrete denoising and regression does not come from the readout.

\begin{table}[htb]
\centering
\caption{Cell type accuracy of spatial generation when types are read from latents by $k$-NN classifiers. The code-to-type readout is in Table~\ref{tab:ablation}.}
\label{tab:latentknn}
\footnotesize
\begin{tabular}{l ccc}
\toprule
Readout & Latent regression & Latent regression (Gaussian) & \textsc{Sudoku} \\
\midrule
$k$-NN, $k=1$         & 0.249 & 0.243 & 0.276 \\
$k$-NN, $k=10$        & 0.294 & 0.290 & 0.336 \\
$k$-NN, $k=50$        & 0.333 & 0.333 & 0.373 \\
$k$-NN, $k=200$       & 0.357 & 0.356 & 0.378 \\
$k$-NN, $k=1{,}000$   & 0.362 & 0.357 & 0.389 \\
$k$-NN, $k=5{,}000$   & 0.360 & 0.368 & 0.397 \\
$k$-NN, $k=10{,}000$  & 0.365 & 0.371 & 0.403 \\
\bottomrule
\end{tabular}
\end{table}

The variant without positional encoding is pre-trained from scratch with the same recipe as \textsc{Sudoku}, fine-tuned in the same way, and evaluated on the same 972 questions.

\clearpage
\section{\texorpdfstring{Additional niche retrieval results}{Additional niche retrieval results}}
\label{app:retrieval-extra}

\paragraph{Chance level.} For a pair of an NOI and a target section, the chance level is the fraction of cells on the target section that carry the organ of the NOI, which is the AUPRC of a random ranking. Table~\ref{tab:chance} gives the chance level $c$ of each organ, averaged over its pairs, and the normalized score $(\mathrm{AUPRC}-c)/(1-c)$, computed from the AUPRC of Table~\ref{tab:retrieval} and the chance level of the organ. The chance level varies from 0.038 to 0.152 across organs, and after normalization \textsc{Sudoku} remains the best method.

\begin{table}[htb]
\centering
\caption{Chance level and chance-normalized AUPRC on HESTA, per organ. \textbf{Bold}: best.}
\label{tab:chance}
\footnotesize
\begin{tabular}{l c ccc}
\toprule
          &              & \multicolumn{3}{c}{Normalized AUPRC} \\
\cmidrule(lr){3-5}
Organ     & Chance level & \textsc{Sudoku} & QueST & GeST \\
\midrule
Brain     & 0.152 & \textbf{0.430} & 0.399 & 0.347 \\
Liver     & 0.076 & \textbf{0.926} & 0.780 & 0.834 \\
Spinal    & 0.135 & \textbf{0.425} & 0.341 & 0.301 \\
Cartilage & 0.151 & \textbf{0.245} & 0.237 & 0.193 \\
Ganglion  & 0.043 & \textbf{0.231} & 0.109 & 0.217 \\
Muscle    & 0.070 & \textbf{0.384} & 0.223 & 0.366 \\
Heart     & 0.038 & \textbf{0.748} & 0.699 & 0.643 \\
Jaw       & 0.061 & 0.155 & 0.069 & \textbf{0.204} \\
\bottomrule
\end{tabular}
\end{table}

\paragraph{Query window size.} For each NOI, we grow the query window from its center through the tissue of the same organ in steps of 250 cells, from 250 to 2{,}000 cells, and keep the NOIs whose organ holds at least 2{,}000 cells, 84{,}064 pairs in all. Table~\ref{tab:window} compares \textsc{Sudoku} with the two strongest baselines of Table~\ref{tab:retrieval}, GeST and QueST. \textsc{Sudoku} performs best at every window size, and its margin over the best baseline widens gradually with the window size, from 4.8\% at 250 cells to 8.9\% at 2{,}000 cells. The score of \textsc{Sudoku} at 1{,}000 cells differs slightly from the 0.503 of Table~\ref{tab:ablation}, since each window here holds exactly 1{,}000 cells, while the benchmark NOIs have a median of about 1{,}000 cells.

\begin{table}[htb]
\centering
\caption{Niche retrieval on HESTA with query windows of growing size, mean AUPRC over the 84{,}064 pairs; gain: relative gain of \textsc{Sudoku} over the best baseline.}
\label{tab:window}
\footnotesize
\setlength{\tabcolsep}{4pt}
\begin{tabular}{l cccccccc}
\toprule
Query window (cells) & 250   & 500   & 750   & 1{,}000 & 1{,}250 & 1{,}500 & 1{,}750 & 2{,}000 \\
\midrule
\textsc{Sudoku}      & 0.501 & 0.501 & 0.504 & 0.509   & 0.510   & 0.512   & 0.513   & 0.512   \\
GeST                 & 0.478 & 0.479 & 0.473 & 0.479   & 0.474   & 0.471   & 0.473   & 0.470   \\
QueST                & 0.447 & 0.450 & 0.449 & 0.451   & 0.452   & 0.452   & 0.449   & 0.451   \\
\midrule
Gain over the best baseline & +4.8\% & +4.6\% & +6.6\% & +6.3\% & +7.6\% & +8.7\% & +8.5\% & +8.9\% \\
\bottomrule
\end{tabular}
\end{table}

\paragraph{Variation across seeds.} In Table~\ref{tab:retrieval}, the values of \textsc{Sudoku} and of the two strongest baselines, QueST and GeST, are means over three runs with seeds 42, 43 and 44; the other baselines were run once due to limited computational resources. Table~\ref{tab:seeds} gives the standard deviations for \textsc{Sudoku} and the best baseline of each column. On HESTA, \textsc{Sudoku} leads on seven of the eight organs, and on six of them the intervals of one standard deviation around the two means do not overlap. On DLPFC, \textsc{Sudoku} leads on five of the seven classes and ties QueST on L3L4, but only the margins on L4L5 and L5L6WM exceed the variation across seeds, and small margins such as L5L6 lie within it. Over the 15 columns of Table~\ref{tab:retrieval}, \textsc{Sudoku} leads on 12, and on 8 of them by more than the variation across seeds.

\begin{table}[htb]
\centering
\caption{Mean $\pm$ standard deviation over seeds 42, 43 and 44 for \textsc{Sudoku} and the best baseline of each column of Table~\ref{tab:retrieval}.}
\label{tab:seeds}
\footnotesize
\begin{tabular}{l cc l}
\toprule
Column of Table~\ref{tab:retrieval} & \textsc{Sudoku} & Best baseline & Baseline \\
\midrule
\multicolumn{4}{l}{\emph{HESTA, AUPRC}} \\
Brain     & 0.517 $\pm$ 0.007 & 0.490 $\pm$ 0.002 & QueST \\
Liver     & 0.932 $\pm$ 0.005 & 0.847 $\pm$ 0.017 & GeST  \\
Spinal    & 0.503 $\pm$ 0.001 & 0.430 $\pm$ 0.010 & QueST \\
Cartilage & 0.359 $\pm$ 0.009 & 0.352 $\pm$ 0.004 & QueST \\
Ganglion  & 0.264 $\pm$ 0.005 & 0.251 $\pm$ 0.006 & GeST  \\
Muscle    & 0.427 $\pm$ 0.006 & 0.410 $\pm$ 0.002 & GeST  \\
Heart     & 0.758 $\pm$ 0.012 & 0.710 $\pm$ 0.007 & QueST \\
Jaw       & 0.207 $\pm$ 0.009 & 0.253 $\pm$ 0.012 & GeST  \\
\midrule
\multicolumn{4}{l}{\emph{DLPFC, PCC}} \\
L3L4      & 0.566 $\pm$ 0.002 & 0.566 $\pm$ 0.004 & QueST \\
L4L5      & 0.698 $\pm$ 0.009 & 0.673 $\pm$ 0.010 & QueST \\
L5L6      & 0.765 $\pm$ 0.005 & 0.762 $\pm$ 0.003 & QueST \\
L6WM      & 0.751 $\pm$ 0.009 & 0.746 $\pm$ 0.011 & QueST \\
L3L4L5    & 0.604 $\pm$ 0.006 & 0.628 $\pm$ 0.001 & QueST \\
L4L5L6    & 0.723 $\pm$ 0.002 & 0.720 $\pm$ 0.004 & QueST \\
L5L6WM    & 0.705 $\pm$ 0.005 & 0.696 $\pm$ 0.002 & QueST \\
\bottomrule
\end{tabular}
\end{table}

\clearpage
\section{\texorpdfstring{Additional spatial generation results}{Additional spatial generation results}}
\label{app:generation-extra}

\paragraph{More baselines.} We add three baselines, scored with the same hard accuracy on the same 972 questions, with the same windows and visible cells: $k$-NN as in Table~\ref{tab:generation} with $k=1,5,50,100,200,300$ and $400$, where larger $k$ is not included because each window has 500 cells, of which at least 20\% are masked, so at most 400 cells are visible; distance-weighted $k$-NN with inverse-distance weights and the same values of $k$; and the composition baseline, which assigns every masked cell the most frequent type among the visible cells of the window, the same as $k$-NN with all visible cells ($k=400$). On the same questions \textsc{Sudoku} reaches 0.401 and GeST 0.362 (Table~\ref{tab:knn}), so \textsc{Sudoku} remains more accurate than all the baselines.

\begin{table}[htb]
\centering
\caption{Spatial generation with $k$-NN baselines over a wide range of $k$, cell-type accuracy on the 972 questions. \textsc{Sudoku} reaches 0.401 and GeST 0.362 on the same questions.}
\label{tab:knn}
\footnotesize
\begin{tabular}{r cc}
\toprule
$k$ & $k$-NN & Distance-weighted $k$-NN \\
\midrule
1   & 0.334 & 0.334 \\
5   & 0.373 & 0.374 \\
50  & 0.372 & 0.376 \\
100 & 0.369 & 0.372 \\
200 & 0.371 & 0.380 \\
300 & 0.384 & 0.385 \\
400 & 0.382 & 0.379 \\
\bottomrule
\end{tabular}
\end{table}

\paragraph{Difficulty tiers and variation across seeds.} We repeat the fine-tuning of \textsc{Sudoku} and GeST with three random seeds. Table~\ref{tab:tiers} gives the accuracy over all questions and in the medium and hard tiers, with the standard deviation across the three runs; the distance-weighted 300-NN, the strongest $k$-NN baseline of Table~\ref{tab:knn}, involves no training or sampling, so its results do not vary across runs. \textsc{Sudoku} stays ahead of the baselines in the medium and hard tiers, which contain the windows with more mixed cell types.

\begin{table}[htb]
\centering
\caption{Spatial generation accuracy over all questions and in the medium and hard tiers; $\pm$: standard deviation across three runs with different seeds.}
\label{tab:tiers}
\footnotesize
\begin{tabular}{l ccc}
\toprule
Method                   & All               & Medium            & Hard              \\
\midrule
\textsc{Sudoku}          & 0.401 $\pm$ 0.003 & 0.379 $\pm$ 0.007 & 0.273 $\pm$ 0.010 \\
GeST                     & 0.362 $\pm$ 0.009 & 0.306 $\pm$ 0.011 & 0.253 $\pm$ 0.008 \\
Distance-weighted 300-NN & 0.385             & 0.333             & 0.258             \\
\bottomrule
\end{tabular}
\end{table}

\paragraph{Most probable type.} In Table~\ref{tab:generation}, \textsc{Sudoku} and GeST assign each masked cell the type of its most probable code (Appendix~\ref{app:benchmarks}). Since each cell type is covered by many codes, this readout may favor a type concentrated on a few codes over a more frequent type spread across many. We therefore also take the most probable type: we sum the predicted probabilities of all codes of each type, in the distribution with which each cell is committed, and take the most probable type. Table~\ref{tab:typereadout} compares the two readouts. With the most probable type, the variant without positional encoding reaches 0.372, close to the $k$-NN baselines, and \textsc{Sudoku} reaches 0.417. The readout changes \textsc{Sudoku} much less than the variant without positional encoding ($+0.016$ vs.\ $+0.055$), as expected from its narrower predicted distribution. The perplexity of the predicted type distribution of each masked cell, which can be read as the effective number of cell types over which the prediction is spread, averages 3.4 for \textsc{Sudoku} and 6.4 for the variant without positional encoding. We keep the most probable code in Table~\ref{tab:generation} because it evaluates what \textsc{Sudoku} generates: each masked cell receives a concrete cell state, and its cell type is read from this state. The most probable type instead summarizes the predicted distribution for one particular set of labels and does not correspond to a generated cell.

\begin{table}[htb]
\centering
\caption{Spatial generation accuracy with two readouts of the cell type. The most probable code is the readout of Table~\ref{tab:generation}.}
\label{tab:typereadout}
\footnotesize
\begin{tabular}{l cc}
\toprule
Method & Most probable code & Most probable type \\
\midrule
$k$-NN, $k=400$                          & n/a   & 0.382 \\
Distance-weighted $k$-NN, $k=300$        & n/a   & 0.385 \\
GeST                                     & 0.362 & 0.373 \\
\textsc{Sudoku} without positional encoding & 0.317 & 0.372 \\
\textsc{Sudoku}                          & 0.401 & 0.417 \\
\bottomrule
\end{tabular}
\end{table}

\clearpage
\section{\texorpdfstring{Zero-shot results}{Zero-shot results}}
\label{app:zeroshot}

Table~\ref{tab:zeroshot} compares \textsc{Sudoku} without fine-tuning, with the pre-trained weights only, with the best zero-shot baseline, that is, the best baseline not fine-tuned on the benchmark sections, and with fine-tuned \textsc{Sudoku}. Pre-training alone already makes \textsc{Sudoku} the best model on HESTA and multi-scale clustering, and fine-tuning helps most where the target data differ from the pre-training corpus. On niche retrieval and multi-scale clustering, zero-shot \textsc{Sudoku} is far above the best zero-shot baseline; on HESTA and multi-scale clustering it already outperforms every baseline on average, including the fine-tuned and task-specific ones (0.486 vs 0.447 for GeST on HESTA, 0.408 vs 0.295 for STAGATE on clustering), and fine-tuning adds 0.010 and 0.044. Fine-tuning helps most on DLPFC (+0.197), the only benchmark measured at spot level, while the pre-training corpus is at single-cell resolution. On spatial generation, zero-shot \textsc{Sudoku} (0.355) is below the best $k$-NN variant (0.385, distance-weighted 300-NN), but fine-tuning on unlabeled sections of the same cohorts brings it above all baselines (0.401).

\begin{table}[htb]
\centering
\caption{\textsc{Sudoku} without fine-tuning, compared with the best zero-shot baseline and with fine-tuned \textsc{Sudoku}. DW: distance-weighted.}
\label{tab:zeroshot}
\footnotesize
\setlength{\tabcolsep}{4pt}
\begin{tabular}{l l c c c}
\toprule
Task                   & Metric                       & \begin{tabular}[b]{@{}c@{}}\textsc{Sudoku}\\zero-shot\end{tabular} & \begin{tabular}[b]{@{}c@{}}Best zero-shot\\baseline\end{tabular} & \begin{tabular}[b]{@{}c@{}}\textsc{Sudoku}\\fine-tuned\end{tabular} \\
\midrule
Niche retrieval, HESTA & AUPRC, mean of 8 organs      & 0.486 & 0.303 (scGPT-spatial) & 0.496 \\
Niche retrieval, DLPFC & PCC, mean of 7 classes       & 0.490 & 0.223 (TERRA)         & 0.687 \\
Multi-scale clustering & ARI, mean of 3 levels        & 0.408 & 0.255 (scGPT-spatial) & 0.452 \\
Spatial generation     & Type accuracy, all questions & 0.355 & 0.385 (DW 300-NN)     & 0.401 \\
\bottomrule
\end{tabular}
\end{table}

\clearpage
\section{\texorpdfstring{Notation}{Notation}}
\label{app:notation}

Table~\ref{tab:notation} lists the symbols of the main text.

\begin{center}
\makeatletter\def\@captype{table}\makeatother
\caption{Notation of the main text, grouped by the section where each symbol first appears. A letter used for two quantities is listed once for each.}
\label{tab:notation}
\footnotesize
\renewcommand{\arraystretch}{1.0}
\begin{tabular}{l p{0.78\linewidth}}
\toprule
Symbol & Meaning \\
\midrule
\multicolumn{2}{l}{\emph{Section 1}} \\
$X$, $P$ & cell states and coordinates of a set of spatially neighboring cells \\
$M$, $\bar M$ & hidden subset of these cells and its complement \\
\midrule
\multicolumn{2}{l}{\emph{Section 3}} \\
$S$, $n$ & a tissue section and its number of cells \\
$i$, $j$, $k$ & cell indices \\
$\mathbf{p}_i\in\mathbb{R}^2$ & physical coordinate of cell $i$ \\
$\mathbf{x}_i\in\mathbb{R}^G$, $G$ & expression vector of cell $i$ over $G$ genes \\
$N_i$, $r$ & neighborhood of cell $i$, the cells within radius $r$ of it \\
\midrule
\multicolumn{2}{l}{\emph{Section 3.1}} \\
$f_\varphi$, $g_\varphi$ & encoder and decoder of the tokenizer, with parameters $\varphi$ \\
$\mathbf{z}_i$ & 128-dimensional tokenizer latent of cell $i$ \\
$\mathbf{e}^{(1)}$, $\mathbf{e}^{(2)}$, $K$ & first- and second-level codebooks, each of $K=256$ code vectors \\
$c^{(1)}_i$, $c^{(2)}_i$, $c_i$ & first- and second-level codes of cell $i$ and the joint code $c_i=(c^{(1)}_i,c^{(2)}_i)$ \\
$\mathrm{NN}(\cdot;\mathbf{e})$ & index of the nearest code vector in codebook $\mathbf{e}$ \\
$\mathbf{z}_{q,i}$ & quantized latent of cell $i$, the sum of its two code vectors \\
$x_{ig}$ & count of gene $g$ in cell $i$ \\
$\boldsymbol{\mu}_i$, $\mathbf{r}_i$ & mean and dispersion of the negative-binomial distribution of each gene of cell $i$ \\
$s_i$ & total count of cell $i$ \\
$\mathrm{NB}$ & negative-binomial likelihood \\
$\mathcal{L}_{\mathrm{tok}}$, $\beta$, $\mathrm{sg}$ & tokenizer loss, commitment weight and stop-gradient \\
\midrule
\multicolumn{2}{l}{\emph{Section 3.2}} \\
$t$, $\epsilon$ & masking rate of a neighborhood, drawn from $U(\epsilon,1)$ with $\epsilon=0.001$ \\
$m_j$ & mask indicator of cell $j$ \\
$M$, $V$ & masked and visible cells of a neighborhood \\
$[\mathrm{M}]$ & mask symbol \\
$\mathbf{W}$, $\mathbf{u}$ & linear projection of a visible latent and learned mask vector \\
$\mathbf{h}^0_j$, $\mathbf{h}_j$ & layer-0 input and final hidden state of cell $j$ \\
$s$, $d_s$ & index of a section and its median nearest-neighbor spacing \\
$\hat{\mathbf{p}}_j$ & coordinate of cell $j$ divided by $d_s$ \\
$D$ & head dimension of attention \\
$\omega_m$, $m$ & rotary frequency of the $m$-th pair of dimensions \\
$\theta$ & base of the rotary frequencies, $\theta=10$ \\
$\mathbf{R}(\hat{\mathbf{p}})$, $\mathbf{R}_x$, $\mathbf{R}_y$ & rotary rotation and its blocks for the $x$ and $y$ components of $\hat{\mathbf{p}}$ \\
$\mathbf{q}_j$, $\mathbf{k}_k$, $a_{jk}$ & query of cell $j$, key of cell $k$ and their attention score \\
$p_\theta$ & spatial transformer, giving the distribution of the codes of a masked cell \\
$\ell$ & code level, 1 or 2 \\
$\mathbf{z}_V$, $\mathbf{p}_N$ & latents of the visible cells and coordinates of all cells in the neighborhood \\
$\mathcal{L}(\theta)$ & pre-training loss of the spatial transformer \\
\midrule
\multicolumn{2}{l}{\emph{Section 3.3}} \\
$T$, $s$ & number of decoding steps and index of a step \\
$M_s$ & cells still masked at step $s$ \\
$\tau$ & sampling temperature \\
$\hat c^{(\ell)}_j$, $\hat c_j$ & drawn level-$\ell$ code and drawn joint code of cell $j$ \\
$g_{jk}$, $k$ & Gumbel noise for code $k$ of cell $j$ \\
$\sigma_j$ & confidence of cell $j$ \\
$k_s$ & number of cells committed at step $s$ \\
$\mathbf{z}_q(\hat c_j)$ & codebook reconstruction of the drawn codes of cell $j$ \\
$\bar{\mathbf{h}}_i$ & neighborhood representation of cell $i$ \\
\midrule
\multicolumn{2}{l}{\emph{Section 4}} \\
$k$ & number of nearest visible cells in the $k$-NN baselines \\
\bottomrule
\end{tabular}
\end{center}

\clearpage
\section{\texorpdfstring{Additional immune-recruitment results}{Additional immune-recruitment results}}
\label{app:spac-extra}

\paragraph{Direct prediction.} We add a direct prediction baseline: a ridge regression model from the mean expression of a clone to its immune recruitment, fitted on the clones of the training sections and scored with the metrics of Section~\ref{sec:spac}. Even on immune recruitment, the one quantity it is trained for, the direct prediction is clearly worse than \textsc{Sudoku} (Table~\ref{tab:direct}).

\begin{table}[htb]
\centering
\caption{Immune-recruitment prediction by \textsc{Sudoku} and by a direct predictor.}
\label{tab:direct}
\footnotesize
\begin{tabular}{l cc}
\toprule
Method            & Spearman, mean of three sections & Median rank error \\
\midrule
\textsc{Sudoku}   & 0.77 & 0.12 \\
Direct prediction & 0.62 & 0.24 \\
\bottomrule
\end{tabular}
\end{table}

\paragraph{Spatial readouts.} Beyond recruitment, the generated band shows where the immune cells are and which cells sit next to the clone, as Figure~\ref{fig:spac}c illustrates for one clone. We score three such quantities on all 125 clones, from the same generated cells as Figure~\ref{fig:spac} and with no retraining. For which side of the clone holds more immune cells, the band is cut into four sectors around the clone, and each sector takes its share of immune cells minus that of the whole band. For which cell types lie within 25\,$\mu$m of the clone, each of 12 cell types takes the share of its cells within 25\,$\mu$m of the clone. For which cell types sit next to each other, each of 67 pairs of cell types takes its count of neighboring cells relative to a random arrangement of the same cells. Table~\ref{tab:readouts} gives the agreement of each quantity with the measurement. \textsc{Sudoku} agrees best on all three, and answers them from the generation it has already made, whereas a direct predictor would need its own target and its own model for each.

\begin{table}[htb]
\centering
\caption{Agreement of three spatial readouts of the generated band with the measurement, over the 125 clones.}
\label{tab:readouts}
\footnotesize
\setlength{\tabcolsep}{4pt}
\begin{tabular}{>{\raggedright\arraybackslash}p{4.3cm} >{\raggedright\arraybackslash}p{3.9cm} ccc}
\toprule
Question about the band around a clone & Agreement with measurement & \textsc{Sudoku} & GeST & \begin{tabular}[b]{@{}c@{}}Expression\\interpolation\end{tabular} \\
\midrule
Which side holds more immune cells     & Pearson over four sectors                     & 0.61 & 0.52 & 0.22 \\
Which cell types lie within 25\,$\mu$m of the clone & Spearman over 12 cell types & 0.81 & 0.46 & 0.32 \\
Which cell types sit next to each other & Pearson over 67 type pairs, against shuffling & 0.74 & 0.39 & 0.29 \\
\bottomrule
\end{tabular}
\end{table}

\clearpage
\section{\texorpdfstring{How the training objective shapes the representation}{How the training objective shapes the representation}}
\label{app:readout-theory}

Representations are read out under full observation (Section~\ref{sec:inference}), whereas the loss of Equation~\ref{eq:loss} is computed on masked cells only. This appendix explains how the objective nevertheless determines the representation, and what the objective of Equation~\ref{eq:loss} contributes compared with a fixed masking rate. In brief, the representation has to reflect how the cells of a neighborhood are arranged, and it is read out under full observation, so its accuracy depends on how well the model predicts one masked cell from all the other cells. Learning this prediction needs two kinds of training context. In the first, most cells are visible, as at readout. In the second, the near cells are masked, which teaches the model to use distant cells. A high fixed masking rate rarely produces the first, a low rate rarely produces the second, and an intermediate rate produces each of them far less often than our objective, which trains every amount of visible context and covers both. We derive this in seven steps:

\begin{enumerate}
\setlength{\itemsep}{1pt}\setlength{\parskip}{0pt}
\item[1.] \emph{What the representation must contain.} The representation of a neighborhood is the average of the hidden states of its cells. For it to reflect more than which cells are present, the state of each cell must also encode which spatial domain the cell belongs to.
\item[2.] \emph{Prediction and domains.} Predicting a masked cell from the cells around it is inferring which domain occupies its position.
\item[3.] \emph{From prediction to representation.} Under full observation, the domain inferred for a cell combines the cell's own state with this same inference from the other cells. The accuracy of the representation therefore depends on how well the model predicts one masked cell from all the others.
\item[4.] \emph{Measurement on the trained model.} That the network stores this inference in the state of a visible cell is an assumption. On the trained model, the state of a visible cell contains most of what the rest of the neighborhood predicts for it.
\item[5.] \emph{Contexts trained by each objective.} Our objective trains every amount of visible context with equal weight, while a fixed masking rate trains only one. At a high rate, a predicted cell rarely has most of its neighbors visible, although this is the context in which representations are read out. At a low rate, the near cells of a predicted cell are rarely all masked, so the model seldom has to learn from distant cells. At an intermediate rate, both contexts occur, but each far less often than under our objective.
\item[6.] \emph{Statistical efficiency.} With this equal weighting, our objective is also the likelihood of the whole neighborhood, so asymptotically it estimates the model at least as precisely as any fixed masking rate.
\item[7.] \emph{Expected gain.} Compared with a low fixed masking rate, we therefore expect about the same representation inside homogeneous tissue and a gain where the local evidence is ambiguous, in agreement with the small difference on organ-level retrieval in Table~\ref{tab:ablation}.
\end{enumerate}

\paragraph{Notation.} Symbols of the main text keep the meaning listed in Table~\ref{tab:notation}, except $n$, $c_j$, $S$, $r$, $\sigma$ and $\ell_j$, which this appendix redefines in the table below. The symbols below are used in more than one step, and all others are defined where they appear.

\begin{center}
\footnotesize
\renewcommand{\arraystretch}{1.0}
\begin{tabular}{l p{0.78\linewidth}}
\toprule
Symbol & Meaning \\
\midrule
$N$, $n$ & a neighborhood and its number of cells \\
$d_j$, $S$ & domain of cell $j$, one of $S$ domains \\
$p(\mathbf{z}\mid d)$ & distribution of the latent $\mathbf{z}$ of a cell in domain $d$ \\
$c_j$, $E$ & first-level code of cell $j$, and the $K\times S$ matrix whose column $d$ is the distribution $p(c\mid d)$ of this code in domain $d$ \\
$\pi(N)$, $\bar\pi$ & domain composition of the neighborhood, and its posterior mean given all cells \\
$\mathbf{b}_j$, $\mathbf{b}^{\mathrm{loo}}_j$ & posterior of the domain of cell $j$ given all cells, and given all cells but $j$ \\
$q$, $\hat q$ & optimal prediction of a masked cell, and the prediction of the network; a hat marks a quantity of the network \\
$\eta_{\mathrm{loo}}$ & excess loss of the prediction when one cell is masked and all others are visible \\
$B$, $\sigma$ & fixed linear map applied to hidden states, and smallest singular value of $E$ \\
$\ell_j(V)$ & loss on cell $j$ when the visible set is $V$ \\
$t_0$ & a fixed masking rate \\
$k$ & number of nearest cells of the predicted cell that are considered \\
$r$ & largest overlap between the latent distributions of two domains \\
\bottomrule
\end{tabular}
\end{center}

\paragraph{Setting.} We describe where a cell sits by the composition of its immediate surroundings, and the arrangement of a neighborhood by these local compositions taken over all of its cells. Accordingly, we model tissue as a set of spatial domains, each a region defined by its cell state composition, that is, by its own distribution of cell states, such as tumor with T cells mixed in, tumor without T cells, or the stroma outside a tumor. Fix a neighborhood $N$ of $n$ cells with coordinates $\mathbf{p}_N$; every distribution below is conditioned on $\mathbf{p}_N$. Each cell $j$ belongs to a domain $d_j\in\{1,\dots,S\}$, and a neighborhood can span several domains. The domain of a cell gives the composition of its surroundings, and the domains of all cells of a neighborhood together give its arrangement. The joint distribution of the domains $d_N$ of the cells is arbitrary, so that it determines the contiguity, order and thickness of domains. Given $d_N$, the cell states are independent, with $\mathbf{z}_j\sim p(\mathbf{z}\mid d_j)$ and therefore $c_j=Q(\mathbf{z}_j)\sim p(c\mid d_j)$, where $Q$ is the quantization of the tokenizer. Two domains may share cell types and differ only in their proportions, and measurement noise is part of $p(\mathbf{z}\mid d)$. The labels used downstream, such as organ or layer, are functions of $d_N$, and we take the domain composition $\pi(N)=\frac1n\sum_{j\in N}\mathbf{e}_{d_j}$, with $\mathbf{e}_d$ the $d$-th unit vector, as the quantity a neighborhood representation should reflect. Its posterior mean given all cells is $\bar\pi=\frac1n\sum_{j}\mathbf{b}_j$, where $\mathbf{b}_j=P(d_j\mid\mathbf{z}_N)$ is the posterior of the domain of cell $j$.

\paragraph{Step 1. What the representation must contain.} Write $\mathbf{h}_j=\Phi(\mathbf{z}_j;\mathcal{C}_j)$ for the final hidden state of cell $j$ under full observation, where $\mathcal{C}_j$ collects the latents and relative positions of the other cells, and $\bar{\mathbf{h}}=\frac1n\sum_j\mathbf{h}_j$ for the neighborhood representation. (a) If $\Phi$ ignores $\mathcal{C}_j$, or if the network does not read the coordinates, $\bar{\mathbf{h}}$ is a function of the multiset $\{\mathbf{z}_k\}_{k\in N}$ alone, that is, of the composition of the neighborhood. (b) If the context enters additively, $\Phi(\mathbf{z}_j;\mathcal{C}_j)=\psi(\mathbf{z}_j)+\sum_{k\ne j}\kappa(\mathbf{p}_k-\mathbf{p}_j)\,\varphi(\mathbf{z}_k)$, exchanging the order of summation gives $\bar{\mathbf{h}}=\frac1n\sum_k\psi(\mathbf{z}_k)+\sum_k w_k\,\varphi(\mathbf{z}_k)$ with $w_k=\frac1n\sum_{j\ne k}\kappa(\mathbf{p}_k-\mathbf{p}_j)$, which is still a composition, weighted by position. Which cells lie next to which can therefore enter the averaged representation only through an interaction between the state of a cell and its context. The posterior $\mathbf{b}_j$ is such a quantity, and its average is $\bar\pi$. The variant without positional encoding in Table~\ref{tab:ablation} is an instance of (a).

\paragraph{Step 2. Prediction and domains.} For any visible set $V$, any masked cell $j\notin V$ and any cell $i$,
\begin{equation}
q(c_j\mid\mathbf{z}_V)=\sum_{d}p(c_j\mid d)\,P(d_j=d\mid\mathbf{z}_V),\qquad
P(d_i\mid\mathbf{z}_N)\;\propto\;p(\mathbf{z}_i\mid d_i)\,P(d_i\mid\mathbf{z}_{N\setminus i}).
\label{eq:fusion}
\end{equation}
The first identity says that the optimal prediction of a masked cell is a mixture of the code distributions of the domains, weighted by the posterior of the domain at its position. If the columns of $E$ are linearly independent, with smallest singular value $\sigma>0$, this posterior is a linear function of the prediction, $P(d_j\mid\mathbf{z}_V)=E^{+}q(\cdot\mid\mathbf{z}_V)$ with $E^{+}$ the pseudo-inverse, as in the analysis of masked language models by \citet{wei2021}.

\emph{Proof.} Both identities follow from the independence of the cell states given the domains. Summing $P(d_{N\setminus j}\mid d_j)\,p(c_j\mid d_j)\,P(\mathbf{z}_V\mid d_V)$ over the domains of the other cells gives $P(c_j,\mathbf{z}_V\mid d_j)=p(c_j\mid d_j)\,P(\mathbf{z}_V\mid d_j)$, which yields the first identity by the law of total probability, and the same factorization with $\mathbf{z}_i$ in place of $c_j$ and $V=N\setminus i$ yields the second by Bayes' rule.

\paragraph{Step 3. From prediction to representation.} The second identity of Equation~\ref{eq:fusion} says that the posterior of the domain of a visible cell under full observation is the likelihood of its own state multiplied by the posterior inferred from the rest of the neighborhood, that is, by the inference made when this cell alone is masked. Let $\hat q(\cdot\mid\mathbf{z}_{N\setminus i})$ be the prediction of the network for cell $i$ when only this cell is masked, and let
\begin{equation}
\eta_{\mathrm{loo}}=\mathbb{E}\,\frac1n\sum_{i\in N}\mathrm{KL}\big(q(\cdot\mid\mathbf{z}_{N\setminus i})\,\big\|\,\hat q(\cdot\mid\mathbf{z}_{N\setminus i})\big)
\end{equation}
be its excess loss in this context. We make two assumptions on the network. (R1) The prediction lies in the model family, $\hat q(\cdot\mid\mathbf{z}_{N\setminus i})=E\,\hat{\mathbf{b}}^{\mathrm{loo}}_i$ for a probability vector $\hat{\mathbf{b}}^{\mathrm{loo}}_i$. (R2) Under full observation the final hidden state encodes the fused posterior linearly: there is a fixed matrix $B$ with $B\mathbf{h}_i=\hat{\mathbf{b}}_i$, where $\hat{\mathbf{b}}_i\propto p(\mathbf{z}_i\mid\cdot)\odot\hat{\mathbf{b}}^{\mathrm{loo}}_i$ is normalized to sum to one. Then
\begin{equation}
\mathbb{E}\,\big\|B\bar{\mathbf{h}}-\bar\pi\big\|_1\;\le\;\frac{2\sqrt{2S}}{\sigma}\,\sqrt{\eta_{\mathrm{loo}}}.
\label{eq:readout-bound}
\end{equation}
The constant of the bound grows with $1/\sigma$.

\emph{Proof.} Write $\mathbf{b}^{\mathrm{loo}}_i=P(d_i\mid\mathbf{z}_{N\setminus i})$. By (R1), the first identity of Equation~\ref{eq:fusion} and Pinsker's inequality, $\|\hat{\mathbf{b}}^{\mathrm{loo}}_i-\mathbf{b}^{\mathrm{loo}}_i\|_1\le\sqrt{S}\,\|\hat{\mathbf{b}}^{\mathrm{loo}}_i-\mathbf{b}^{\mathrm{loo}}_i\|_2\le\frac{\sqrt{S}}{\sigma}\|\hat q-q\|_1\le\frac{\sqrt{S}}{\sigma}\sqrt{2\,\mathrm{KL}(q\,\|\,\hat q)}$. Next, let $P(d,\mathbf{z})=b^{\mathrm{loo}}_{i,d}\,p(\mathbf{z}\mid d)$ and $\hat P(d,\mathbf{z})=\hat b^{\mathrm{loo}}_{i,d}\,p(\mathbf{z}\mid d)$, whose conditionals given $\mathbf{z}=\mathbf{z}_i$ are $\mathbf{b}_i$ and $\hat{\mathbf{b}}_i$ by the second identity. Averaging over $\mathbf{z}_i$ drawn from the marginal of $P$ and using the triangle inequality, $\mathbb{E}_{\mathbf{z}_i}\|\hat{\mathbf{b}}_i-\mathbf{b}_i\|_1\le\|P-\hat P\|_1+\|P_{\mathbf{z}}-\hat P_{\mathbf{z}}\|_1\le2\,\|\hat{\mathbf{b}}^{\mathrm{loo}}_i-\mathbf{b}^{\mathrm{loo}}_i\|_1$. By (R2), $\|B\bar{\mathbf{h}}-\bar\pi\|_1\le\frac1n\sum_i\|\hat{\mathbf{b}}_i-\mathbf{b}_i\|_1$, and Jensen's inequality moves the expectation inside the square root.

\begin{figure}[htb]
\centering
\includegraphics[width=\linewidth]{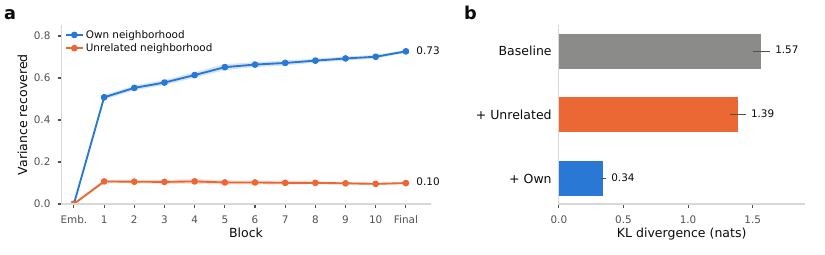}
\caption{Test of assumption (R2) on the \textsc{Sudoku} model used for HESTA retrieval. (a) Share of the variance of the final hidden state at a cell when it alone is masked, beyond what the baseline of the cell's input and coordinate-only features explains, that a linear probe recovers from the state of the visible cell after each block, for the cell in its own neighborhood and for the same cell moved to an unrelated neighborhood. (b) Divergence between the distribution decoded from the probe and the actual prediction with the cell masked, for the baseline alone and with the final state of the visible cell added. Bands and error bars are the standard deviation over the 5 folds.}
\label{fig:readout-probe}
\end{figure}

\paragraph{Step 4. Measurement on the trained model.} We test assumption (R2) on the \textsc{Sudoku} model used for HESTA retrieval, and find that the final hidden state of a visible cell contains, in a linearly readable form, most of what the rest of the neighborhood predicts for it. Assumption (R2) does not follow from the training objective, since the loss is computed on masked cells only. Masked and visible cells are processed by the same network and differ only in their own input, and the states of visible cells at the input of every block receive gradient because masked cells attend to them, but the update of the last block at visible cells is read by no masked cell and receives none. The assumption is the same for our objective and for a fixed masking rate. From each of the 23 HESTA sections we draw 100 neighborhoods and take the central cell and 9 other cells of each, 23{,}000 cells in all. For each cell we record its hidden states under full observation and, from a second forward pass in which only this cell is masked, the final hidden state at its position, from which the two heads compute the prediction. A ridge regression then predicts this state, with 5-fold cross-validation across sections. The baseline uses the latent and the codes of the cell together with features computed from coordinates alone, namely the size of the neighborhood, the distance of the cell to its center, local cell counts and nearest-neighbor distances. Adding the final hidden state of the cell under full observation explains 73\% of the variance that the baseline leaves unexplained, with a standard deviation of 0.7\% across folds, and lowers the divergence between the distribution decoded from the regression and the actual prediction with the cell masked from 1.57 to 0.34 nats, summed over the two code levels. As a control, we place the latent of the same cell at a random position of another neighborhood and use its final hidden state there, which has the same input of the cell but an unrelated context. It explains 10\% (Figure~\ref{fig:readout-probe}). The input embedding, which contains no context, explains 0.1\%. The share is 69\% before the last block and 73\% after it and the final normalization, so the update that receives no gradient does not remove this information.

\paragraph{Step 5. Contexts trained by each objective.} Masks are drawn independently of the data, so every masking scheme has the same optimal predictor $q$, and the schemes differ in how the excess loss is weighted over contexts. For any network, let $\ell_j(V)$ be the loss on cell $j$ when the visible set is $V$. Cell $j$ is masked with probability $t$ independently of the other cells, so the weight $1/t$ cancels this probability:
\begin{equation}
\mathbb{E}_{M\sim\mathrm{Bern}(t)}\,\frac1t\sum_{j\in M}\ell_j(N\setminus M)=\sum_{j\in N}\mathbb{E}_{V\sim t}\,\ell_j(V),
\label{eq:pertarget}
\end{equation}
where under $V\sim t$ every cell other than $j$ is visible independently with probability $1-t$. Equation~\ref{eq:loss} is thus a sum over cells in which every cell is predicted once, at a masking rate $t$ that is uniform. A fixed masking rate $t_0$ keeps a single value of $t$, and dropping the factor $1/t$ gives $t$ the density $2t$. Table~\ref{tab:context-weights} lists the resulting probability of three contexts per predicted cell. The entries are Beta integrals, and the lower limit $\epsilon$ multiplies the first entry of the first row by $(1-\epsilon)^{n-1}$.

\begin{table}[htb]
\centering
\caption{Probability of three contexts per predicted cell under each masking scheme, for a neighborhood of $n$ cells and the $k$ nearest cells of the predicted cell.}
\label{tab:context-weights}
\footnotesize
\begin{tabular}{l ccc}
\toprule
Context of the predicted cell & Equation~\ref{eq:loss} & Without the factor $1/t$ & Fixed masking rate $t_0$ \\
\midrule
All other cells visible       & $1/n$     & $2/(n(n+1))$     & $(1-t_0)^{n-1}$ \\
The $k$ nearest cells visible & $1/(k+1)$ & $2/((k+1)(k+2))$ & $(1-t_0)^{k}$ \\
The $k$ nearest cells masked  & $1/(k+1)$ & $2/(k+2)$        & $t_0^{\,k}$ \\
\bottomrule
\end{tabular}
\end{table}

\emph{The context of the readout.} A high fixed masking rate gives almost no weight to the context with one cell masked, on which the readout depends by step 3. Since the excess loss is an average of non-negative divergences, a model with excess loss $\delta$ under a scheme that gives the context with one cell masked probability $\gamma$ satisfies $\eta_{\mathrm{loo}}\le\delta/\gamma$, and a predictor that errs only in this context attains equality. With Equation~\ref{eq:readout-bound}, the readout error is controlled by $n\delta$ under Equation~\ref{eq:loss}, by $\frac{n(n+1)}{2}\delta$ without the factor $1/t$, and by $(1-t_0)^{-(n-1)}\delta$ at a fixed masking rate. For $n=100$ the three probabilities are $10^{-2}$, $2\times10^{-4}$ and $10^{-7}$ at $t_0=0.15$, or $10^{-60}$ at $t_0=0.75$. Both parts of Equation~\ref{eq:loss} matter for this context. Letting $t$ reach zero includes it in the training distribution, and the factor $1/t$ keeps it from being diluted by neighborhoods with many masked cells. At the scale of the $k$ nearest cells, a fixed masking rate of 0.15 gives the all-visible context at least the weight of Equation~\ref{eq:loss} up to $k=18$ and exponentially less beyond, and a fixed masking rate of 0.75 gives it less from $k=1$ on. The latter matches the drop of linear-probing accuracy, from 73.5\% to 59.6\%, that \citet{mae} report when an encoder trained with mask tokens at a masking rate of 0.75 is applied to unmasked images.

\emph{Distant cells.} Masking many cells trains how the posterior depends on distant cells, which receive training signal only when the near cells are masked or inconclusive. Suppose the $k$ nearest cells of cell $j$ share its domain, and let $r<1$ be the largest overlap $\int\sqrt{p(\mathbf{z}\mid d)\,p(\mathbf{z}\mid d')}\,\mathrm{d}\mathbf{z}$ between two domains. If $v$ of these $k$ cells are visible, the reduction of the loss that the remaining visible cells can bring is at most $I(c_j;\mathbf{z}_{\mathrm{far}}\mid\mathbf{z}_{\mathrm{near}})\le H(d_j\mid\mathbf{z}_{\mathrm{near}})\le(S-1)\ln2\cdot r^{v}$. Averaged over $v$, the factor $r^v$ becomes $(t_0+(1-t_0)r)^k$ at a fixed masking rate and $\frac{1-r^{k+1}}{(k+1)(1-r)}$ under Equation~\ref{eq:loss}; for pure domains, $r=0$, these are the entries $t_0^{\,k}$ and $1/(k+1)$ of Table~\ref{tab:context-weights}. For $k\ge2$ no fixed masking rate reaches the weight of Equation~\ref{eq:loss} in both the second and the third row of the table, since $(1-t_0)^k\,t_0^{\,k}\le4^{-k}<(k+1)^{-2}$.

\emph{Proof.} The first inequality holds because $c_j$ depends on the other cells only through $d_j$. The second uses $H(\mathbf{p})=\sum_a p_a\ln\big(1+\sum_{b\ne a}p_b/p_a\big)\le\sum_{a<b}(p_a+p_b)\,h\big(\tfrac{p_a}{p_a+p_b}\big)$, the bound $h(x)\le2\ln2\sqrt{x(1-x)}$ on the binary entropy $h$, the fact that the expectation of $\sqrt{p_ap_b}$ over $v$ independent observations is $\sqrt{\nu_a\nu_b}$ times the $v$-th power of the overlap, where $\nu$ is the prior of $d_j$, and $\sum_{a<b}\sqrt{\nu_a\nu_b}\le\frac{S-1}{2}$.

\paragraph{Step 6. Statistical efficiency.} Let $\{p_\theta\}$ be a regular parametric family of distributions over the tokens of a neighborhood, and suppose that the conditionals computed by the network are those of $p_\theta$. With $\epsilon\to0$, Equation~\ref{eq:pertarget} makes the number of visible cells uniform on $\{0,\dots,n-1\}$ for every predicted cell, so Equation~\ref{eq:loss} averaged over masks is the expectation, over a uniformly random order $\rho$ of the cells, of $\sum_{i}-\log p_\theta(c_{\rho_i}\mid\mathbf{z}_{\rho_{<i}})$. For consistent conditionals every order gives the same value $-\log p_\theta(\mathbf{x}_N)+\sum_j\log q_{\mathrm{tok}}(\mathbf{z}_j\mid c_j)$ by the chain rule, so minimizing Equation~\ref{eq:loss} is maximum likelihood; for a general network it is the bound of Appendix~\ref{app:bound}. A fixed masking rate instead minimizes a sum of conditional log-likelihoods of single cells, a composite likelihood, in which the predictions made within one neighborhood are correlated but enter as if they were independent. Its gradient $g$ is an unbiased estimating function. With $s$ the score of $p_\theta$, $\mathcal{I}=\mathbb{E}[ss^\top]$ and $A=\mathbb{E}[gs^\top]$, the asymptotic covariance of its minimizer satisfies $A^{-1}\,\mathbb{E}[gg^\top]\,A^{-\top}\succeq\mathcal{I}^{-1}$, the covariance of maximum likelihood \citep{godambe1960}, with equality only if $g$ is a linear function of $s$. Every parameter of the tissue model, and every quantity computed from the parameters, such as the prediction with one cell masked that the readout fuses, is thus estimated with an asymptotic variance no larger under Equation~\ref{eq:loss} than at any fixed masking rate. As an illustration, take two arrangements $Y=\pm1$ of a neighborhood of 13 cells, a central cell that indicates $Y$ correctly with probability 0.75, and twelve flanking cells that, in half of the neighborhoods, each indicate $Y$ correctly with probability 0.95 and are otherwise uninformative, and estimate the prior of $Y$. The asymptotic mean squared error of the probability that two neighborhoods share an arrangement, computed from their posteriors, is 4.96, 1.08 and 1.04 times the Cram\'er--Rao bound at $t_0=0.15$, at $t_0=0.75$ and for uniform $t$ without the factor $1/t$, and it attains the bound for Equation~\ref{eq:loss}. These comparisons take the average over masks. With a single mask per neighborhood the factor $1/t$ adds variance, and the order of the schemes can change in small examples.

\paragraph{Step 7. Expected gain.} The overlap $r$ of step 5 bounds what distant cells can change in the readout. For a cell $i$ whose $k$ nearest cells share its domain,
\begin{equation}
\mathbb{E}\,\big\|P(d_i\mid\mathbf{z}_N)-P(d_i\mid\mathbf{z}_i,\mathbf{z}_{\mathrm{near}})\big\|_1\;\le\;4\,\mathbb{E}\big[1-\max_dP(d_i=d\mid\mathbf{z}_i,\mathbf{z}_{\mathrm{near}})\big]\;\le\;2(S-1)\,r^{\,k+1}.
\label{eq:far-bound}
\end{equation}
The gain of our objective over a low fixed masking rate is concentrated at borders, in domains one or two cells thick and between domains of similar composition, and should be small for neighborhoods inside one organ. This agrees with the small difference between \textsc{Sudoku} and the fixed masking rate of 0.15 on organ-level retrieval in Table~\ref{tab:ablation}.

\emph{Proof.} Let $\mathbf{b}$ be the posterior from the cell and its near cells, $a=1-\max_db_d$, $f_d$ the likelihood of the distant cells under domain $d$ and $m=\sum_db_df_d$. The expected change of the posterior is $\sum_db_d\|f_d-m\|_1$. For the most probable domain $\|f_d-m\|_1\le2a$, and the other domains have total mass $a$ with $\|f_d-m\|_1\le2$ each, which gives $4a$. The second inequality follows from $1-\max_dp_d\le\sum_{d<d'}\sqrt{p_dp_{d'}}$ and the same expectation over $k+1$ independent observations.

\paragraph{Scope.} Equations~\ref{eq:fusion}, \ref{eq:pertarget} and \ref{eq:far-bound} and Table~\ref{tab:context-weights} need only the tissue model or hold for any network. Equation~\ref{eq:readout-bound} additionally needs (R1) and (R2), of which (R2) is supported by the measurement of step 4. The statement on efficiency is asymptotic and assumes a well-specified model. We do not claim that our objective outperforms every fixed masking rate on every task, since any distribution of $t$ with mass near zero trains the context of the readout, and a single well-chosen masking rate can match one scale. Our objective is the one weighting that covers every scale at once and coincides with the likelihood, so no masking rate needs to be tuned across tissues whose domains, overlaps and neighborhood sizes differ.

\end{document}